\documentclass[11pt]{article}
\pdfoutput=1
\usepackage{enumerate}
\usepackage{pdfsync}
\usepackage[OT1]{fontenc}
\usepackage{natbib}
\usepackage[usenames]{color}
\usepackage{uclaml}
\usepackage[colorlinks,
            linkcolor=red,
            anchorcolor=blue,
            citecolor=blue
            ]{hyperref}
\usepackage{fullpage}
\usepackage{hyperref}
\usepackage[protrusion=true,expansion=true]{microtype}
\usepackage{pbox}
\usepackage{setspace}
\usepackage{tabularx}
\usepackage{float}
\usepackage{wrapfig,lipsum}
\usepackage{enumitem}
\usepackage{microtype}
\usepackage{graphicx}
\usepackage{subfigure}
\usepackage{enumerate}
\usepackage{enumitem}
\usepackage{pgfplots}
\usetikzlibrary{arrows,shapes,snakes,automata,backgrounds,petri}
\usepackage{booktabs} % for professional tables

\allowdisplaybreaks
\usepackage{enumitem}
\usepackage{algpseudocode}
\makeatletter
\newcommand*{\rom}[1]{\expandafter\@slowromancap\romannumeral #1@}
\makeatother
\title{\huge Locally Differentially Private Reinforcement Learning for Linear Mixture Markov Decision Processes}

\author
{
    Chonghua Liao\thanks{Department of Automation, Tsinghua University, Beijing, China, 100084; e-mail: {\tt lch18@mails.tsinghua.edu.cn}} 
    ~~~and~~~
	Jiafan He\thanks{Department of Computer Science, University of California, Los Angeles, CA 90095, USA; e-mail: {\tt jiafanhe19@ucla.edu}} 
	~~~and~~~
	Quanquan Gu\thanks{Department of Computer Science, University of California, Los Angeles, CA 90095, USA; e-mail: {\tt qgu@cs.ucla.edu}}
}

\date{}

\begin{document}
\maketitle

\begin{abstract}
Reinforcement learning (RL) algorithms can be used to provide personalized services, which rely on users' private and sensitive data. To protect the users' privacy, privacy-preserving RL algorithms are in demand. In this paper, we study RL with linear function approximation and local differential privacy (LDP) guarantees. We propose a novel $(\varepsilon, \delta)$-LDP algorithm for learning a class of Markov decision processes (MDPs) dubbed linear mixture MDPs, and obtains an $\tilde{\mathcal{O}}( d^{5/4}H^{7/4}T^{3/4}\left(\log(1/\delta)\right)^{1/4}\sqrt{1/\varepsilon})$ regret, where $d$ is the dimension of feature mapping, $H$ is the length of the planning horizon, and $T$ is the number of interactions with the environment. 
We also prove a lower bound $\Omega(dH\sqrt{T}/\left(e^{\varepsilon}(e^{\varepsilon}-1)\right))$ for learning linear mixture MDPs under $\varepsilon$-LDP constraint. Experiments on synthetic datasets verify the effectiveness of our algorithm. To the best of our knowledge, this is the first provable privacy-preserving RL algorithm with linear function approximation. 
\end{abstract}

\section{Introduction}
Reinforcement learning (RL) algorithms have been studied extensively in the past decade. When the state and action spaces are large or even infinite, traditional tabular RL algorithms \citep[e.g.,][]{watkins1989learning, jaksch2010near, azar2017minimax} become computationally inefficient or even intractable. To overcome this limitation, modern RL algorithms with function approximation are proposed, which often make use of feature mappings to map states and actions to a low-dimensional space. This greatly expands the application scope of RL. While RL can provide personalized service such as online recommendation and personalized advertisement, existing algorithms rely heavily on user's sensitive data. Recently, how to protect sensitive information has become a central research problem in machine learning. For example, in online recommendation systems, users want accurate recommendation from the online shopping website to improve their shopping experience while preserving their personal information such as demographic information and purchase history. 

Differential privacy (DP) is a solid and highly successful notion of algorithmic privacy introduced by \citet{dwork2006calibrating}, which indicates that changing or removing a single data point will have little influence on any observable output. However, DP is vulnerable to membership inference attacks \citep{shokri2017membership} and has the risk of data leakage. To overcome the limitation of DP, a stronger notion of privacy, \textit{local differential privacy} (LDP), was introduced by \citet{kasiviswanathan2011can,duchi2013local}. Under LDP, users will send privatized data to the server and each individual user maintains its own sensitive data. The server, on the other hand, is totally agnostic about the sensitive data. 

(Local) differential privacy has been extensively studied in multi-armed bandit problems, which can be seen as a special case of MDPs with unit planing horizon and without state transition. Nevertheless, \citet{shariff2018differentially} proved that the standard DP is incompatible in the contextual bandit setting, which will yield a linear regret bound in the worst case. Therefore, they studied a relaxed version of DP named \textit{jointly differentially private} (JDP), %This definition is introduced by \citet{shariff2018differentially}, and it
which basically requires that changing one data point in the collection of information from previous users will not have too much influence on the decision of the future users. %\qg{add a reference for JDP}. 
Recently, LDP has attracted increasing attention in multi-armed bandits. %\qg{add a few papers}. 
\citet{gajane2018corrupt} are the first to study LDP in stochastic multi-armed bandits (MABs).  \citet{chen2020locally} studied combinatorial bandits with LDP guarantees. \citet{zheng2020locally} studied both MABs and contextual bandits, and proposed a locally differentially private algorithm for contextual linear bandits. However, differentially private RL is much less studied compared with bandits, even though MDPs are more powerful since state transition is rather common in real applications. For example, a user may click a link provided by the recommendation system to visit a related webpage, which can be viewed as state transition. In tabular RL, \citet{vietri2020private} proposed a $\varepsilon$-JDP algorithm and proved an $\tilde{\cO}(\sqrt{H^4SAT} + SAH^3(S + H)/\varepsilon)$ regret, where $H$ is the length of the planning horizon, $S$ and $A$ are the number of states and actions respectively, $K$ is the number of episodes, and $T = KH$ is the number of interactions with the MDP. %\qg{explain H, S, A, T, blabla}.
Recently, \citet{garcelon2020local} designed the first LDP tabular RL algorithm with an $\tilde{\cO}(\max\cbr{H^{3/2}S^2A\sqrt{T}/\varepsilon, HS\sqrt{AT}})$ regret. However, as we mentioned before, tabular RL algorithms suffer from computational inefficiency when applied to large state and action spaces. Therefore, a natural question arises:

 %To understand how LDP can protect the user's sensitive data in RL, let us take online recommendation as an example. At time $t$, a user is presented with a set of items (states). The platform will decide in what order these items should be presented to the user (actions), and it will observe the time a user spend watching the items (rewards). %In linear mixture MDPs \citep{jia2020model, ayoub2020model,zhou2020provably} (See Definition~\ref{definMDP} for more details.), the predefined feature embedding $\bphi(s^\prime, s,a)$ contains sensitive information of states and actions, and the value function $V(s)$ contains information of rewards. Both should be protected before sending to the server.

\begin{center}
    \textit{Can we design a privacy-preserving RL algorithm with linear function approximation while maintaining the statistical utility of data?}
\end{center}

In this paper, we answer this question affirmatively. More specifically, we propose a locally differentially private algorithm for learning linear mixture MDPs \citep{jia2020model, ayoub2020model,zhou2020provably} (See Definition~\ref{definMDP} for more details.), where the transition probability kernel is a linear function of a predefined $d$-dimensional feature mapping over state-action-next-state triple. The key idea is to inject Gaussian 
noises into the sensitive information in the UCRL-VTR backbone, and the main challenge is how to balance the tradeoff between the Gaussian perturbations for privacy preservation and the utility to learn the optimal policy.  %\qg{need 1-2 sentences to describe the design of your algorithm. e.g., what does your algorithm look like.}%We follow the idea of LSVI-UCB algorithm in \citet{jin2020provably} and provide a novel privacy-preserving algorithm named LDP-UCRL-VTR. 

Our contributions are summarized as follows.
\begin{itemize}[leftmargin = *]
    \item We propose a novel algorithm named LDP-UCRL-VTR to learn the optimal value function while protecting the sensitive information. We show that our algorithm guarantees $\rbr{\varepsilon, \delta}$-LDP and enjoys an $\tilde{\cO}( d^{5/4}H^{7/4}T^{3/4}\rbr{\log(1/\delta)}^{1/4}\sqrt{1/\varepsilon})$ regret bound, where $T$ is the number of rounds and $H$ is the length of episodes. To our knowledge, this is the first locally differentially private algorithm for RL with linear function approximation.
    \item %Based on the lower bound $\Omega(d\sqrt{T})$ in \citet{lattimore2020bandit} for non-private linear bandit algorithms, 
    We prove a $\Omega(dH\sqrt{T}/\rbr{e^{\varepsilon}(e^{\varepsilon}-1)})$ lower bound for learning linear mixture MDPs under  $\varepsilon$-LDP constraints. Our lower bound suggests that the aforementioned upper bound might be improvable in some parameters (i.e., $d,H,T$). %\qg{compare the lower bound with upper bound}. 
    % \cl{dependence on d}
    As a byproduct, our lower bound also implies $\Omega(d\sqrt{T}/\rbr{e^{\varepsilon}(e^{\varepsilon}-1)})$ lower bound for $\varepsilon$-LDP contextual linear bandits. This suggests that the algorithms proposed in \citet{zheng2020locally} might be improvable as well.
\end{itemize}
\paragraph{Notation} We use lower case letters to denote scalars, lower and upper case bold letters to denote vectors and matrices. For a vector $\xb\in \RR^d$ , we denote by $\|\xb\|_1$ the Manhattan norm and denote by $\|\xb\|_2$ the Euclidean norm. For a semi-positive definite matrix $\bSigma$ and any vector $\xb$, we define $\nbr{\xb}_{\bSigma}:=\nbr{\bSigma^{1/2}\xb}_2 = \sqrt{\xb^{\top} \bSigma \xb}$. $\mathds{1}(\cdot)$ is used to denote the indicator function. For any positive integer $n$, we denote by $[n]$ the set $\cbr{1, \dots, n}$. For any finite set $\cA$, we denote by $\abr{\cA}$ the cardinality of $\cA$. We also use the standard $\cO$ and $\Omega$ notations, and the notation $\tilde{\cO}$ is used to hide logarithmic factors. We denote $D_{1:h} = \cbr{D_1, \dots, D_h}$. For two distributions $p$ and $p^{\prime}$, we define the Kullback–Leibler divergence (KL-divergence) between $p$ and $p^{\prime}$ as follows: 
\[\text{KL}(p, p^{\prime}) = \int p(\zb)\log \frac{p(\zb)}{p^{\prime}(\zb)}\ud \zb.\]
%\qg{define KL divergence here and update the notation for KL throughout the paper}
\section{Related Work}
\textbf{Reinforcement Learning with Linear Function Approximation} Recently, there have been many advances in RL with function approximation, especially in the linear case. \citet{jin2020provably} considered linear MDPs where the transition probability and the reward are both linear functions with respect to a feature mapping $\bphi: \cS\times \cA \rightarrow \RR^d$, and proposed an efficient algorithm for linear MDPs with $\tilde{\cO}(\sqrt{d^3H^3T})$ regret.  \citet{yang2019sample} assumed the probabilistic transition model has a linear structure. They also assumed that the features of all state-action pairs can be written as a convex combination of the anchoring features. \citet{wang2019optimism} designed a statistically and computationally efficient algorithm with generalized linear function approximation, which attains an  $\tilde{\cO}(H\sqrt{d^3T})$ regret bound.  \citet{zanette2020learning} proposed RLSVI algorithm with $\tilde{\cO}(d^2\sqrt{H^4T})$ regret bound under the linear MDPs assumption. \citet{jiang2017contextual} studied a larger class of MDPs with low Bellman rank and proposed an OLIVE algorithm with polynomial sample complexity. Another line of work considered linear mixture MDPs (a.k.a., linear kernel MDPs) \citep{jia2020model, ayoub2020model,zhou2020provably}, which assumes the transition probability function is parameterized as a linear function of a given feature mapping on a triplet $\bpsi:\cS \times \cA \times \cS \to \RR^d$. \citet{jia2020model} proposed a model-based RL algorithm, UCRL-VTR, which attains an $\tilde{\cO}(d\sqrt{H^3T})$ regret bound. \citet{ayoub2020model} considered the same model but with general function approximation, and proved a regret bound depending on Eluder dimension \citep{russo2013eluder}. \citet{zhou2020nearly} proposed an improved algorithm which achieves the nearly minimax optimal regret. \citet{he2020logarithmic} showed that logarithmic regret is attainable for learning both linear MDPs and linear mixture MDPs. 

\noindent\textbf{Differentially Private Bandits} The notion of \textit{differential privacy} (DP) was first introduced in \citet{dwork2006calibrating} and has been extensively studied in both MAB and  contextual linear bandits.  \citet{basu2019differential} unified different privacy definitions and proved an  $\Omega(\sqrt{KT}/\rbr{e^{\varepsilon}(e^{\varepsilon}-1)})$ regret lower  bound for locally differentially private MAB algorithms, where $K$ is the number of arms. \citet{shariff2018differentially} derived an impossibility result for learning contextual bandits under DP constraint by showing an $\Omega(T)$ regret lower bound for any $(\varepsilon, \delta)$-DP algorithms. Hence, they considered the relaxed \textit{joint differential privacy} (JDP) and proposed an algorithm based on Lin-UCB \citep{abbasi2011improved} with $\tilde{\cO}(\sqrt{T}/\varepsilon)$ regret while preserving $\varepsilon$-JDP. Recently, a stronger definition of privacy, \textit{local differential privacy} \citep{duchi2013local,kasiviswanathan2011can}, gained increasing interest in bandit problems. Intuitively, LDP ensures that each collected trajectory is differentially private when observed by the  agent, while DP requires the computation on the entire set of trajectories to be DP. \citet{zheng2020locally} proposed an LDP contextual linear bandit algorithm with $\tilde{\cO}(d^{3/4}T^{3/4})$ regret.

\textbf{Differentially Private RL} In RL, \citet{balle2016differentially} is the first to propose a private algorithm for policy evaluation with linear function approximation. In the tabular setting, \citet{vietri2020private} designed a $\varepsilon$-JDP algorithm for regret minimization which attains an $\tilde{\cO}(\sqrt{H^4SAT} + SAH^3(S + H)/\varepsilon)$ regret. Recently, \citet{garcelon2020local} presented an optimistic algorithm with LDP guarantees. Their algorithm enjoys an $\tilde{\cO}(\max\cbr{H^{3/2}S^2A\sqrt{T}/\varepsilon, HS\sqrt{AT}})$ regret upper bound. They also provided a $\tilde{\Omega}(\sqrt{HSAT}/\min\cbr{\exp(\varepsilon)-1,1})$ regret lower bound. However, all these private RL algorithms are in the tabular setting, and private RL algorithms with linear function approximation remain understudied.
\section{Preliminaries}
In this paper, we study locally differentially private RL with linear function approximation for episodic MDPs. In the following, we will introduce the necessary background and definitions.

\subsection{Markov Decision Processes}
\paragraph{Episodic Markov Decision Processes} We consider the setting of an episodic time-inhomogeneous Markov decision process \citep{puterman2014markov}, denoted by a tuple $M = M(\cS, \cA, H, \cbr{r_h}^H_{h=1}, \cbr{\PP_h}_{h=1}^H)$, where $\cS$ is the state space, $\cA$ is the action space, $H$ is the length of each episode (planning horizon), $r_h: \cS\times \cA\rightarrow [0,1]$ is the deterministic reward function and $\PP_h(s'|s,a) $ is the transition probability function which denotes the probability for state $s$ to transfer to state $s'$ given action $a$ at stage $h$. A policy $\pi = \cbr{\pi_h}_{h=1}^H$ is a collection of $H$ functions, where $\pi_h(s)$ denote the action that the agent will take at stage $h$ and state $s$. Moreover, for each $h\in [H]$, we define the value function $V^{\pi}_h: \cS\rightarrow \RR$ that maps state $s$ to the expected value of cumulative rewards received under policy $\pi$ when starting from state $s$ at the $h$-th stage. We also define the action-value function $Q_h^{\pi}: \cS\times \cA\rightarrow \RR$ 
which maps a state-action pair $(s,a)$ to the expected value of cumulative rewards when the agent starts from  state-action pair $(s,a)$ at the $h$-th stage and follows policy $\pi$ afterwards. More specifically, for each state-action pair $(s,a) \in \cS \times \cA$, we have
\begin{align*}
    &Q_h^{\pi}(s,a)=r_h(s,a)+\EE\sbr{\sum_{h^{\prime}=h+1}^H r_{h^{\prime}}\big(s_{h^{\prime}},\pi_{h^{\prime}}(s_h^{\prime})\big)}\,, 
    V_h^{\pi}(s) = Q_h^{\pi}\big(s, \pi_h(s)\big)\,,
\end{align*}
where $s_h = s, a_h = a$ and $s_{h^{\prime}+1}\sim \PP_{h'}(\cdot\given s_{h^{\prime}}, a_{h^{\prime}})$.

For each function $V:\cS\rightarrow \RR$, we further denote $\sbr{\PP_h V}(s,a) = \EE_{s^{\prime}\sim \PP_h(\cdot\given s,a)}V(s^{\prime})$. Using this notation, the Bellman equation with policy $\pi$ can be written as
\begin{align*}
    Q^{\pi}_h(s,a) = r_h(s,a)+\sbr{\PP_{h} V^{\pi}_{h+1}}(s,a),\qquad V_h^{\pi}(s) = Q_h^{\pi}(s, \pi_h(s)), \qquad V_{H+1}^{\pi}(s)=0,
\end{align*}
We define the optimal value function $V_h^*$ as $V_h^*(s) = \max_{\pi}V_h^{\pi}(s)$ and the optimal action-value function $Q_h^*$ as $Q_h^*(s,a) = \max_{\pi}Q_h^{\pi}(s,a)$. With this notation, the Bellman optimality equation can be written as follows
\begin{align*}
    Q^{*}_h(s,a) = r_h(s,a)+\sbr{\PP_{h} V^{*}_{h+1}}(s,a), \qquad V^{*}_{h+1}(s) = \max_{a\in \cA}Q^{*}_h(s,a),\qquad V^{*}_{H+1}(s) = 0\,.
\end{align*}

In the setting of an episodic MDP, an agent aims to learn the optimal policy by interacting with the environment and observing the past information. At the beginning of the $k$-th episode, the agent chooses the policy $\pi_k$ and the adversary picks the initial state $s_1^k$. At each stage $h\in[H]$, the agent observes the state $s_h^k$, chooses an action following the policy $a_h^k=\pi_h^k(s_h^k)$ and observes the next state with $s_{h+1}^k \sim \PP_h(\cdot|s_h^k,a_h^k)$.
The difference between $V_1^{*}(s_1^k)$ and $V_1^{\pi^k}(s_1^k)$ represents the expected regret in the $k$-th episode. Thus, the total regret in first $K$ episodes can be defined as
\[\text{Regret}(K)=\sum_{k=1}^{K} \rbr{V_1^{*}(s_1^k)-V_1^{\pi_k}(s_1^k)}\,.\]
\paragraph{Linear Function Approximation} In this work, we consider a class of MDPs called \textit{linear mixture MDPs} \citep{jia2020model, ayoub2020model,zhou2020provably}, where the transition probability function can be represented as a linear function of a given feature mapping $\bphi(s^\prime, s,a): \cS\times\cA\times\cS\to \RR^d$ satisfying that for any bounded function $V: S \to [0, H]$ and any tuple $(s, a) \in S \times A$, we have
\begin{equation}
    \left\|\boldsymbol{\phi}_{V}(s, a)\right\|_{2} \leq H, \text { where } \boldsymbol{\phi}_{V}(s, a)=\sum_{s^{\prime} \in \mathcal{S}} \boldsymbol{\phi}(s^{\prime} \given s, a) V(s^{\prime})\,.\label{linearMixure}
\end{equation}
Formally, we have the following definition:
\begin{definition}[\citealt{jia2020model, ayoub2020model,zhou2020provably}]\label{definMDP}
An MDP $(\cS, \cA, H, \cbr{r_h}^H_{h=1}, \cbr{\PP_h}_{h=1}^H)$ is an inhomogeneous, episodic bounded linear mixture MDP if there exist vectors $\btheta_h\in \RR^d$ with $\nbr{\btheta^*_h}\le \sqrt{d}$ and a feature map $\bphi: \cS\times \cA \times \cS\rightarrow \RR^d$ satisfying \eqref{linearMixure} such that $\PP_h(s^\prime\given s, a) = \left<\bphi(s^{\prime}\given s, a), \btheta^*_h\right>$ for any state-action-next-state triplet $(s, a, s^\prime) \in S \times A \times S$ and stage $h$.
\end{definition}

Therefore, learning the underlying $\btheta_h^*$ can be regarded as solving a ``linear bandit'' problem (Part V, \citet{lattimore2020bandit}), where the context is $\bphi_{V_{k,h+1}}(s^k_h, a^k_h) \in \RR^d$, and the noise is $V_{k,h+1}(s^k_{h+1}) - [\PP_h V_{k,h+1}](s^k_h, a^k_h)$.

%\CC{comment on Linear MDP}

%\subsection{Local Differential Privacy}

\subsection{(Local) Differential Privacy}\label{sseDP}
In this subsection, we introduce the standard definition of differential privacy \citep{dwork2006calibrating} and local differential privacy \citep{kasiviswanathan2011can, duchi2013local}. We also present the definition of Gaussian mechanism.
\paragraph{Differential Privacy} Differential privacy is a mathematically rigorous notion of data privacy. In our setting, DP considers that the information collected from all the users can be observed and aggregated by a server. It ensures that the algorithm's output renders neighboring inputs indistinguishable. Thus, we formalize the definition as follows:
\begin{definition}[Differential Privacy]\label{definitionDP}
     For any user $k\in [K]$, let $D_k$ be the information sent to a 
privacy-preserving mechanism from user $k$ and the collection of data from all the users can be written as $\cbr{D_{k}}_{1:K}= \{D_1,\dots, D_k, \dots, D_K\}$. For any $\varepsilon\ge0$ and $\delta\ge 0$, a randomized mechanism $\cM$ preserves $\rbr{\varepsilon, \delta}$-differential privacy if for any two neighboring datasets $\cbr{D_{k}}_{1:K},\cbr{D_{k}^{\prime}}_{1:K} \in \cZ$ which only different at one entry, and for any measurable subset $U\in \cZ$, it satisfies
    \[\PP\rbr{\cM\rbr{\cbr{D_{k}}_{1:K}} \in U} \leq e^{\varepsilon}\PP\rbr{\cM\rbr{\cbr{D_{k}^{\prime}}_{1:K}} \in U} +\delta\,, \]
    where $\cbr{D_k}_{1:K} = \{D_1,\dots, D_k, \dots, D_K\}, \cbr{D_k^{\prime}}_{1:K} = \{D_1,\dots, D_{k}^\prime, \dots, D_K\}$.
\end{definition}
\paragraph{Local Differential Privacy}
In online RL, we view each episode $k \in [K]$ as a trajectory associated to a specific user. A natural way to conceive LDP in RL setting is to guarantee that for any user, the information send to the server has been privatized. Thus, LDP ensures that the server is totally agnostic to the sensitive data, and we are going to state the following definition:
\begin{definition}[Local Differential Privacy]\label{definitionLDP}
 For any $\varepsilon\ge0$ and $\delta\ge 0$, a randomized mechanism $\cM$ preserves $\rbr{\varepsilon, \delta}$-local differential privacy if for any two users $u$ and $u^{\prime}$ and their corresponding data $D_{u}, D_{u^{\prime}}\in \cU$, it satisfies:
    \[\PP\rbr{\cM\rbr{D_{u}} \in U} \leq e^{\varepsilon}\PP\rbr{\cM\rbr{D_{u^\prime}} \in U} +\delta,\quad U\in \cU\,. \]
    %where $\cM$ is a privacy preserving mechanism.
\end{definition}
\begin{remark}
The dataset $\cbr{D_k}_{1:K}$ in DP is a collection of information from users $1, \dots, K$, where the subscript indicates the $k$-th user. Post-processing theorem implies that LDP is a more strict notion of privacy than DP. 
\end{remark}

Now we are going to introduce the Gaussian mechanism which is widely used as a privacy-preserving mechanism to ensure DP/LDP property.
% begin{lemma}\label{wishartDP}
%     (Theorem 4.1, \citealt{sheffet2015private}). Fix $\varepsilon \in (0,1) $ and $\delta \in (0,1/e )$. Fix $B>0$. Let $\Ab$ be a $n \times d$-matrix where each row of $\Ab$ has bounded $\ell_2$-norm of $B$. Let $\Nb$ be a matrix sampled from the d-dimensional Wishart distribution with $\nu$-degrees of freedom using the scale matrix $B^2\cdot \Ib_{d \times d}$(i.e., $\Nb \sim \cW_d(B^2\cdot \Ib_{d \times d},\nu)$) for $\nu \geq \left\lfloor d+(28/\varepsilon^2) \cdot \log(4/\delta)\right\rfloor $. Then outputting $\Xb=\Ab^{\top}\Ab +\Nb$ is $(\varepsilon,\delta)$-differentially private.
% \end{lemma}

% Now we introduce the standard definition of the Gaussian mechanism.
\begin{lemma}\label{gaussian mechanism}
    (The Gaussian Mechanism, \citealt{dwork2014algorithmic}). Let $ f : \NN^{\cX} \mapsto \RR^d $ be an arbitrary d-dimensional function (a query), and define its $\ell_2$ sensitivity as $\Delta_2f=\max _{\text{adjacent(x,y)}} \nbr{f(x)-f(y)}_{2} $, where $\text{adjacent}(x,y)$ indicates that $x,y$ are different only at one entry. For any $0\leq \varepsilon \leq 1 $ and $c^2>2\log(1.25/\delta)$, the Gaussian Mechanism with parameter $\sigma \geq c\Delta_2f/\varepsilon$ is $(\varepsilon,\delta)$-differentially private. 
\end{lemma}
\section{Algorithm}\label{sectionAlgorithm}
We propose LDP-UCRL-VTR algorithm as displayed in Algorithm \ref{algorithm}, which can be regarded as a variant of UCRL-VTR algorithm proposed in \citet{jia2020model} with $\rbr{\varepsilon, \delta}$-LDP guarantee. Algorithm \ref{algorithm} takes the privacy parameters $\varepsilon, \delta$ as input (Line \ref{1}). For the first user $k = 1$, we simply have $\bLambda_{1, h} = \bSigma_{1,h} = \lambda \Ib$ and $\hat{\btheta}_{1, h} = \mathbf{0}$ (Line \ref{4}). For local user $k$ and received information $\bLambda_h^k, \ub_h^{k}$, the optimistic estimator of the optimal action-value function is constructed with an additional UCB bonus term (Line \ref{6}),
\begin{align*}
    Q_{k,h}(\cdot, \cdot)&\gets \min\left\{H, r_h(\cdot, \cdot) + \left<\hat{\btheta}_{k,h}, \bphi_{V_{k, h+1}}(\cdot, \cdot)\right> + \beta \nbr{\bSigma_{k,h}^{-1/2}\bphi_{V_{k,h+1}}(\cdot, \cdot)}_2\right\}\,,
\end{align*}
and $\beta$ is specified as $c d^{3/4}(H-h+1)^{3/2}k^{1/4}\log(dT/\alpha)\rbr{\log((H-h+1)/\delta}^{1/4}\sqrt{1/\varepsilon}$, where $c$ is an absolute constant. From the previous sections, we know that learning the underlying $\btheta_h^*$ can be regarded as solving a ``linear bandit'' problem, where the context is $\bphi_{V_{k,h+1}}(s^k_h, a^k_h) \in \RR^d$, and the noise is $V_{k,h+1}(s^k_{h+1}) - [\PP_h V_{k,h+1}](s^k_h, a^k_h)$. Therefore, to estimate $Q_h^*$, it suffices to estimate the vector $\btheta_h^*$ by ridge regression with input $\bphi_{V_{k, h+1}}(s_h^k,a_h^k)$ and output $V_{k,h+1}(s_{h+1}^k)$. In order to implement the ridge regression, the server should collect the information of $\bphi_{V_{k, h+1}}(s_h^k,a_h^k)\bphi_{V_{k, h+1}}(s_h^k,a_h^k)^\top$ and $V_{k,h+1}(s_{h+1}^k)$ from each user $k$ (Line \ref{14}). Thus, we need to add noises to privatize the data before sending these information to the server in order to kept user's information private. In LDP-UCRL-VTR, we attain LDP by adding a symmetric Gaussian matrix and $d$-dimensional Gaussian noise (Line \ref{11} and Line \ref{12}). For simplicity, we denote the original information (without noise) $\Delta \tilde{\bLambda}_h^{k}=\bphi_{V_{k, h+1}}(s_h,a_h)\bphi_{V_{k, h+1}}(s_h,a_h)^\top, \Delta\tilde{\ub}_h^{k}= \bphi_{V_{k, h+1}}(s_h^k,a_h^k)V_{k, h+1}(s_{h+1}^k)$, where $k$ indicates the user and $h$ indicates the stage. Since the input information to the server is kept private by the user, it is easy to show that LDP-UCRL-VTR algorithm satisfies $(\varepsilon, \delta)$-LDP. After receiving the information from user $1$ to user $k$, the server aggregates information $\bLambda_{k,h}, \ub_{k,h}$, and maintains them for $H$ stages separately (Line \ref{18}). Besides, since the Gaussian matrix may not preserve the PSD property, we adapt the idea of shifted regularizer in \citet{shariff2018differentially} and shift this matrix $\bLambda_{k,h}$ by $r\Ib$ to guarantee PSD (Line \ref{19}). We then calculate $\hat{\btheta}_{k+1,h}$ and send $\bSigma_{k+1,h}, \hat{\btheta}_{k+1,h}$ back to $k+1$-th user  in order to get a more precise estimation of $\btheta_h^*$ for better exploration.

\paragraph{Comparison with related algorithms.}
We would like to comment on the difference between our LDP-UCRL-VTR and other related algorithm. The key difference between our LDP-UCRL-VTR and UCRL-VTR \citep{jia2020model}, which is the most related algorithm, is that we add additive noises to the contextual vectors and the optimistic value functions in order to guarantee privacy. Then the server collects privatized information from different users and update $\bLambda, \ub$ for ridge regression. A shifted regularizer designed in \citet{shariff2018differentially} is used to guarantee PSD property of the matrix. It is easy to show that if we add no noise to user's information, our LDP-UCRL-VTR algorithm degenerates to inhomogeneous UCRL-VTR. Another related algorithm is the Contextual Linear Bandits with LDP in \citet{zheng2020locally}, which is an algorithm designed for contextual linear bandits. Setting $H= 1$, our LDP-UCRL-VTR will degenerate to Contextual Linear Bandits with LDP in \citet{zheng2020locally}.

\begin{algorithm}[t!]
    \caption{LDP-UCRL-VTR}
    \label{algorithm}
    \begin{algorithmic}[1]
        \Require privacy parameters $\varepsilon, \delta$, failure probability $\alpha$, parameter $\lambda$
        \State Set $\sigma=4H^3\sqrt{2\log(2.5H/\delta)}/\varepsilon$ \label{1}
        \For{user $k = 1,\dots, K$}
        \State\textbf{For the local user $k$:}
        \State Receive $\{\bSigma_{k, 1},\dots, \bSigma_{k,H},\hat{\btheta}_{k, 1},\dots,\hat{\btheta}_{k, H}\}$ from the server\label{4}
        \For{$h = H, \dots, 1$}
        \State $Q_{k,h}(\cdot, \cdot)\gets \min\left\{H - h 
        + 1, r_h(\cdot, \cdot)+ \left<\hat{\btheta}_{k,h}, \bphi_{V_{k, h+1}}(\cdot, \cdot)\right> +\beta_h \nbr{\bSigma_{k,h}^{-1/2}\bphi_{V_{k,h+1}}(\cdot, \cdot)}_2\right\}$\label{6}
        \State $V_{k,h}(\cdot)\gets \max_{a} Q_{k,h}(\cdot,a).$
        \EndFor
        \State Receive the initial state $s_1^k$
        \For{$h = 1, \dots, H$}
        \State Take action $a_h^k \gets \argmax_{a\in \cA} Q_{k,h}(s_h^k)$, and observe $s_{h+1}^k$\label{10}
        \State Set $\Delta \bLambda_h\gets\bphi_{V_{k, h+1}}(s_h,a_h)\bphi_{V_{k, h+1}}(s_h,a_h)^\top+\Wb_{k,h}$, where $\Wb_{k,h}(i,j) = \Wb_{k,h}(j,i)$ and  $\Wb_{k,h}(i, j)\overset{\text{i.i.d}}{\sim} \mathcal{N}(0, \sigma^2), \forall i \le j$\label{11}
        \State Set $\Delta \ub_{h}\gets\bphi_{V_{k, h+1}}(s_h^k,a_h^k)V_{k, h+1}(s_{h+1}^k)+\bxi_h $, where $\bxi_h \sim \mathcal{N}(\mathbf{0}_d, \sigma^2 \Ib_{d\times d})$\label{12}
        \EndFor
        \State Send $D^k=\{ \Delta \bLambda_1,\dots,\Delta \bLambda_H,\Delta\ub_1,\dots,\Delta\ub_H\}$ to the server\label{14}
        \Statex
        \State\textbf{For the server:}
        \For{$h = 1, \dots, H$}
        \State $\bLambda_{k+1, h} \gets \bLambda_{k,h} + \Delta \bLambda_h, \ub_{k+1,h} \gets \ub_{k,h} + \Delta \ub_h$\label{18}
        \State $\bSigma_{k+1,h} \gets \bLambda_{k+1,h} + r\Ib$\label{19}
        \State $\hat{\btheta}_{k+1,h} \gets \rbr{\bSigma_{k+1,h}}^{-1}\ub_{k+1,h}$
        \EndFor
        \State Send $\{\bSigma_{k+1,1},\dots, \bSigma_{k+1,H},\hat{\btheta}_{k+1, 1},\dots,\hat{\btheta}_{k+1, H}\}$ to user $k+1$
        \EndFor
    \end{algorithmic}
\end{algorithm}

\section{Main Results}

In this section, we provide both privacy and regret guarantees for Algorithm \ref{algorithm}. The detailed proofs of the main results are deferred to the appendix.

\subsection{Privacy Guarantees}

%\qg{Add sentences before or after each theorem.}
%For private linear bandits learning, \citet{shariff2018differentially} and \citet{zheng2020locally} designed a similar framework to protect privacy. The goal is to protect the contextual vector and the corresponding reward. 

Recall that in Algorithm \ref{algorithm}, we use Gaussian mechanism to protect the private information of the contextual vectors and the optimistic value functions. Based on the property of Gaussian mechanism, we can show that our algorithm is $(\varepsilon, \delta)$-LDP. 

\begin{theorem}\label{LDPtheorem}
Algorithm \ref{algorithm} preserves $(\varepsilon,\delta)$-LDP.
\end{theorem}
%\begin{remark}
The privacy analysis relies on the fact that if the information from each user satisfies $(\varepsilon, \delta)$-DP, then the whole algorithm is $(\varepsilon, \delta)$-LDP. %Therefore, the information collected by the server is perturbed by $k$ sources of noise. %Compared with Algorithm 1 in \citet{shariff2018differentially}, which only needs to add $\cO(\log(K))$ perturbations by tree-based algorithm in order to satisfy $(\varepsilon, \delta)$-JDP, our LDP-UCRL-VTR needs to add more noises. 
%\end{remark}

\subsection{Regret Upper Bound}
The following theorem states the regret upper bound of Algorithm \ref{algorithm}. %It shows that by setting $\beta$ and $\lambda$ properly, our LDP-UCRL-VTR enjoys a $\tilde{\cO}(T^{3/4})$ regret upper bound, 
\begin{theorem}\label{mainTheorem}
     For any fixed $\alpha \in (0, 1)$, for any privacy parameters $\varepsilon > 0$ and $\delta>0$, if we set the parameters $\lambda = 1$ and $\beta_h = \tilde{\cO}( d^{3/4}(H-h+1)^{3/2}k^{1/4}\log(dT/\alpha)\rbr{\log((H-h+1)/\delta)}^{1/4}\sqrt{1/\varepsilon})$ for user $k$, with probability at least $1-\alpha$, the total regret of Algorithm \ref{algorithm} in the first $T$ steps is at most $\tilde{O}(d^{5/4}H^{7/4}T^{3/4}\log(dT/\alpha)\rbr{\log(H/\delta)}^{1/4}\sqrt{1/\varepsilon})$, where $T = KH$ is the number of interactions with the MDP.
\end{theorem}
% \begin{proof}
% See Appendix \ref{appenMainTheorem}.
% \end{proof}
\begin{remark}
By setting $\alpha = \delta$, our regret bound can be written as $\tilde{\cO}(d^{5/4}H^{7/4}T^{3/4}\rbr{\log(H/\delta)}^{1/4}\sqrt{1/\varepsilon})$. Compared with UCRL-VTR, which enjoys an upper bound of $\tilde{\cO}(d\sqrt{H^3T})$, our bound suggests that learning the linear mixture MDP under the LDP constraint is inherently harder than learning it non-privately. 
\end{remark}

\subsection{Regret Lower Bounds} \label{lower}
In this subsection, we present a lower bound for learning linear mixture MDPs under the $\varepsilon$-LDP constraint. We follow the idea firstly developed in \citet{zhou2020nearly}, which basically shows that learning a linear mixture MDP is no harder than learning $H/2$ linear bandit problems. As a byproduct, we also derived the regret lower bound for learning $\varepsilon$-LDP contextual linear bandits.

%We adapt the hard-to-learn MDP instance constructed in \citet{zhou2020nearly} with a sightly different $\Delta$ and

In detail, in order to prove the regret lower bound for MDPs under $\varepsilon$-LDP constraint, we first prove the lower bound for learning $\varepsilon$-LDP linear bandit problems. We adapted the proof techniques in \citet[Theorem 24.1]{lattimore2020bandit} and \citet{basu2019differential}. In the non-private setting, the observed history of a contextual bandit algorithm in the first $T$ rounds can be written as $\cH_T = \cbr{\xb_t, y_t}_{t = 1}^T$. Given history $\cH_{t-1}$, the contextual linear bandit algorithm chooses action $\xb_t$, and the reward is generated from a distribution $f_{\btheta}(\cdot | \xb_t)$, which is conditionally independent of the previously observed history. We use $\PP_{\pi,{\btheta}}^T$ to denote the distribution of observed history up to time $T$, which is induced by $\pi$ and $f_{\btheta}$. Hence, we have
\begin{align}
    \PP^T_{\pi, {\btheta}} = \PP_{\pi, {\btheta}}(\cH_T) = \prod^T_{t=1}\pi (\xb_t\given \cH_{t-1})f_{\btheta}(y_t\given \xb_t)\,,\notag
\end{align}
    where $\pi$ is the stochastic policy (the distribution over an action set induced by a bandit algorithm) and $f_{\btheta}(\cdot\given \xb_t)$ is the reward distribution given action $\xb_t$, which is conditionally independent of the previously observed history $\cH_{t-1}$.
    
In the LDP setting, the privacy-preserving mechanism $\cM$ generates the privatized version of the context $\xb_t$, denoted by $\tilde \xb_t = \cM(\xb_t)$, to the contextual linear bandit algorithm. For simplicity, we denote $\cM_{\pi}$ as the distribution (stochastic policy) by imposing a locally differentially private mechanism $\cM$ on the distribution (policy) $\pi$. Also, we use $f^{\cM}_{\btheta}$ to denote the conditional distribution of $\tilde y_t$ parameterized by $\btheta$, where $\tilde y_t$ is the privatized version of $y_t$ obtained by the privacy-preserving mechanism $\cM$. We denote the observed history by $\tilde{\cH}_T:=\cbr{\rbr{\tilde{\xb}_t, \tilde{y}_t}}_{t=1}^{T}$, where $\tilde{\xb}_t$, $\tilde{y}_t$ are the privatized version of contexts and rewards. Similarly, we have\
\begin{align}
    \tilde{\PP}_{\pi, {\btheta}}^T:=\PP_{\pi, {\btheta}}(\tilde{\cH}_T) = \prod_{t=1}^T\cM_{\pi}(\tilde{\xb}_t\given \tilde{\cH}_{t-1}) f^{\cM}_{\btheta}(\tilde{y}_t\given \xb_t)\,.\label{eq:definition of P}
\end{align}

With the formulation above, we proved the following key lemma for $\varepsilon$-LDP contextual linear bandits. 
%We follow  and provide a locally differentially private KL-divergence decomposition lemma.% in Appendix \ref{AppenC}.%, see Lemma \ref{KLdecom}.
\begin{lemma}\label{KLdecom}
    (Locally Differentially Private KL-divergence Decomposition) We denote the reward generated by user $t$ for action $\xb_t$ as $y_t = \xb_t^\top \btheta + \eta_t$, where $\eta_t$ is a zero-mean noise. If the reward generation process is $\varepsilon$-locally differentially private for both the bandits with parameters $\btheta_1$ and $\btheta_2$, we have,
    \begin{align*}
        \text{KL}(\tilde{\PP}_{\pi,{\btheta_1}}^T,\tilde{\PP}_{\pi,{\btheta_2}}^T)&\le 2\min\cbr{4, e^{2\varepsilon}}(e^{\varepsilon}-1)^2 \cdot \sum_{t=1}^T \EE_{\pi,{\btheta_1}}\sbr{\text{KL}(f^{\cM}_{\btheta_1}(\tilde{y}_t\given \xb_t), f^{\cM}_{\btheta_2}(\tilde{y}_t\given \xb_t))},
    \end{align*}
    where $\tilde{\xb}_t$, $\tilde{y}_t$ are the privatized version of contexts and rewards.
\end{lemma}

Lemma \ref{KLdecom} can be seen as an extension of Lemma 3 in \citet{basu2019differential} from multi-armed bandits to contextual linear bandits. 

Equipped with Lemma \ref{KLdecom}, the KL-divergence of privatized history distributions can be decomposed into the distributions of rewards. %We know from Bretagnolle-Huber inequality \citep{bretagnolle1979estimation}, that the probability of an event $\cE$ and its complement $\cE^C$ for two distributions $P$ and $Q$ is lower bounded by $\exp\rbr{-\text{KL}\rbr{P,Q}}$. Thus, we focus on the special reward distribution we specified earlier and construct $d$ pairs of complementary events, where $d$ is the dimension of feature embedding.
%\begin{remark}
We construct a contextual linear bandit with Bernoulli reward. In detail, for an action $\xb_t \in \cA \subseteq \RR^d$, the reward follows a Bernoulli distribution $y_t \sim B(\left<\btheta,\xb_t\right> + \delta)$, where $0\le\delta\le 1/3$. %Nevertheless, in order to utilize this lemma in the hard-to-learn MDPs, we assume that the reward distribution is a Bernoulli distribution. 
We first  derive a regret lower bound of learning contextual bandits under the LDP constraint in the following lemma.
\begin{lemma}[Regret Lower Bound for LDP Contextual Linear Bandits]\label{banditLowerBound} Given an $\varepsilon$-locally differentially private reward generation mechanism with $\varepsilon$ and a time horizon $T$, for any environment with finite variance, the pseudo regret of any algorithm $\pi$ satisfies
\[\text{Regret}(T) \ge \frac{c}{\min\cbr{2, e^{\varepsilon}}(e^{\varepsilon}-1)}d\sqrt{T}\,.\]
\end{lemma}

Since the distribution of rewards will only influence the KL-divergence by an absolute constant, the lower bound we obtained is similar to that in \citet[Theorem 24.1]{lattimore2020bandit}, which assumes that the reward follows a normal distribution.
%\end{remark}

According to the proof of Lemma \ref{banditLowerBound}, the only difference between our hard-to-learn MDP instance and that in \citet{zhou2020nearly} is that we need to specify $\Delta = \sqrt{\delta}/\rbr{\min\cbr{2, e^{\varepsilon}}(e^{\varepsilon}-1)\sqrt{T}}$. %In this case, we only need to give a lower bound of $\varepsilon$-LDP contextual linear bandits. 
We then utilize the hard-to-learn MDPs constructed in \citet{zhou2020nearly} and obtain the following lower bound for learning linear mixture MDPs with $\varepsilon$-LDP guarantee:
\begin{theorem}\label{RLlower}
For any $\varepsilon$-LDP algorithm there exists a linear mixture MDP parameterized by $\bTheta = \rbr{\btheta_1, \dots, \btheta_H}$ such that the expected regret is lower bounded as follows:
\[\EE_{\bTheta}\text{Regret}(M_{\bTheta}, K) \ge \Omega\rbr{\frac{1}{\min\cbr{2,e^{\varepsilon}}(e^{\varepsilon}-1)}dH\sqrt{T}},\]
where $T = KH$ and $\EE_{\bTheta}$ denotes the expectation over the probability distribution generated by the interaction of the algorithm and the MDP.
\end{theorem}
\begin{remark}
Compared with the upper bound $\tilde{\cO}(d^{5/4}H^{7/4}T^{3/4}\rbr{\log(H/\delta)}^{1/4}\sqrt{1/\varepsilon})$ in Theorem~\ref{mainTheorem}, it can be seen that there is a $d^{1/4}T^{1/4}H^{3/4}$ gap between our upper bound and lower bound if treating $\varepsilon$ as a constant. It is unclear if the upper bound and/or the lower bound are not tight.%\CC{comment on the dependence on $T$ and }
\end{remark}
\begin{remark}
Notice that $\varepsilon$-LDP is a special case of $(\varepsilon, \delta)$-LDP, where $\delta = 0$. Thus our lower bound for $\varepsilon$-LDP linear mixture MDPs is also a valid lower bound for learning $(\varepsilon, \delta)$-LDP linear mixture MDPs.
\end{remark}

\begin{figure*}[!ht]
    \centering
    \begin{tikzpicture}%[global scale = 0.5]
    \node[circle,
    minimum width =35pt ,
    minimum height =35pt ,draw=blue!70, fill=blue!20, very thick] (0) at(0,0){$S_1$} ;
    \node[circle,
    minimum width =35pt ,
    minimum height =35pt ,draw=blue!70, fill=blue!20, very thick] (1) at(2.5,0){$S_2$};
    \node[circle,
    minimum width =25pt ,
    minimum height =25pt ,draw=blue!0, fill=blue!0, very thick] (2) at(5,0){$\dots$};
    \node[circle,
    minimum width =35pt ,
    minimum height =35pt ,draw=blue!70, fill=blue!20, very thick] (3) at(7.5,0){$S_{S -1}$};
    \node[circle,
    minimum width =35pt ,
    minimum height =35pt ,draw=blue!70, fill=blue!20, very thick] (4) at(10,0){$S_{S}$};

    \path[->, bend angle=25, shorten >=1pt, semithick] 
        (0) edge[bend left] node[above]{$ap_h$}  (1)
            edge[loop left] node{$1-ap_h$}  (0)
        (1) edge[bend left] node[below]{$(1-ap_h)/2$}  (0)
        edge[bend left] node[above]{$ap_h$}  (2)
        edge[loop above]    node[above]{$(1-ap_h)/2$}  (1)
        (2) edge[bend left]  node[below]{$(1-ap_h)/2$} (1)
        edge[bend left]  node[above]{$ap_h$} (3)
        (3) edge[bend left]  node[below]{$(1-ap_h)/2$} (2)
        edge[bend left] node[above]{$ap_h$} (4)
        edge[loop above]    node[above]{$(1-ap_h)/2$}  (3)
        (4) edge[bend left] node[below]{$(1-ap_h)/2$} (3)
        edge[loop right]  node{$1-ap_h$}  (4);
    \end{tikzpicture}
    \caption{The transition kernel $\PP_h$ of inhomogeneous ``RiverSwim'' MDP instance.}
    \label{fig:mdp}
\end{figure*}
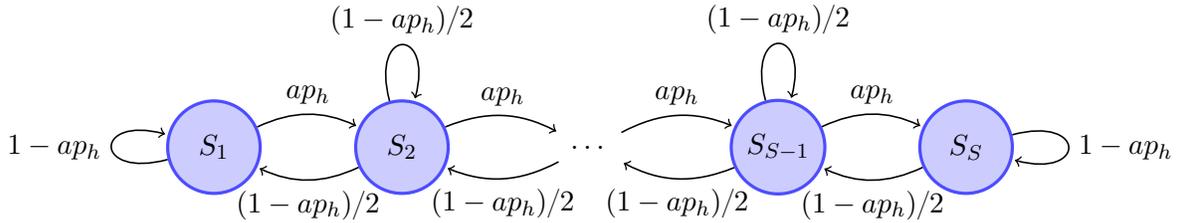

\begin{figure*}[!ht]
    \subfigure[$S=3, A=2, H = 6, d=18$, homogeneous]{%
      \includegraphics[width=0.44\linewidth]{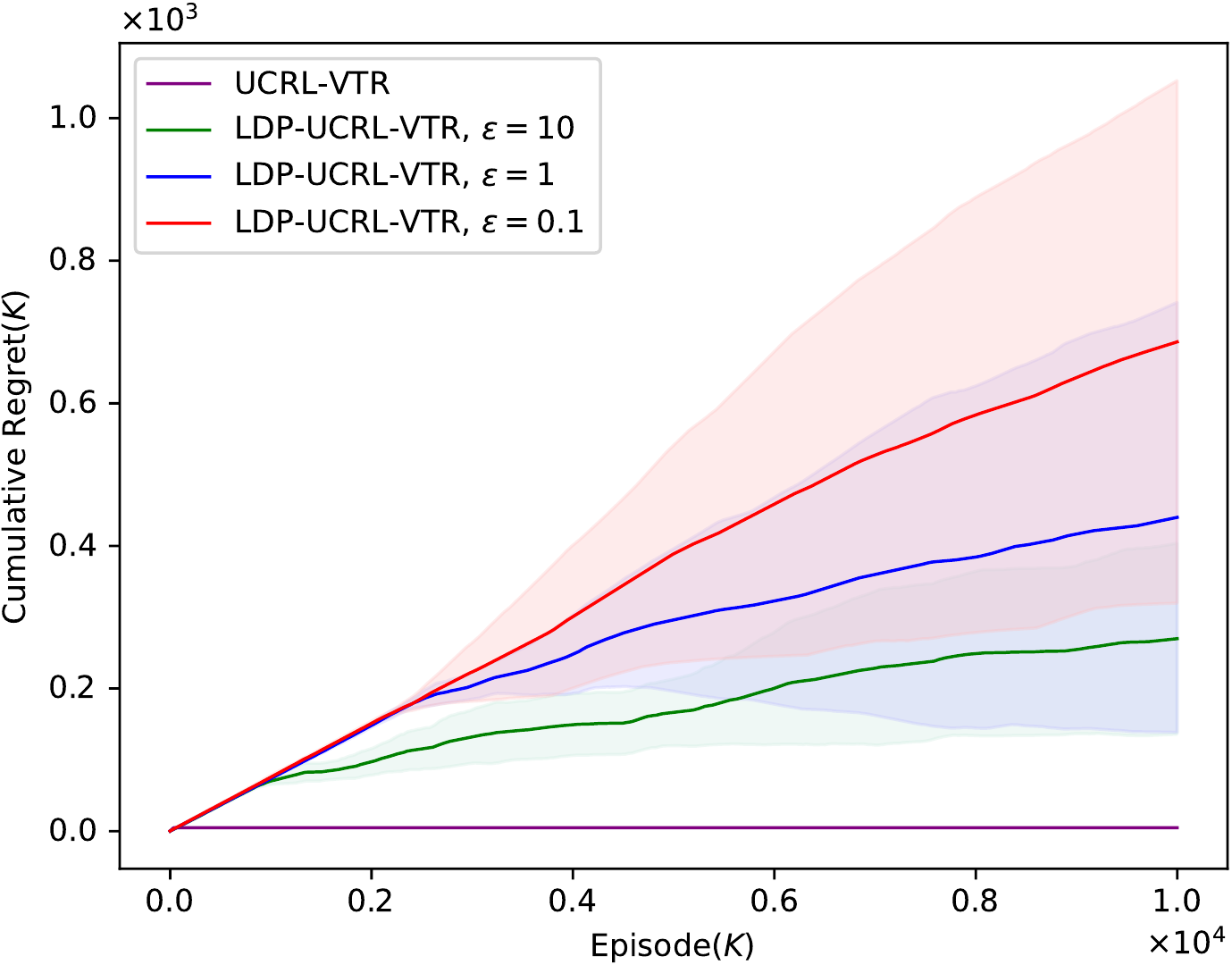}
    }
    \hfill
    \subfigure[$S=5, A = 2, H = 10, d=50$, homogeneous]{%
      \includegraphics[width=0.44\linewidth]{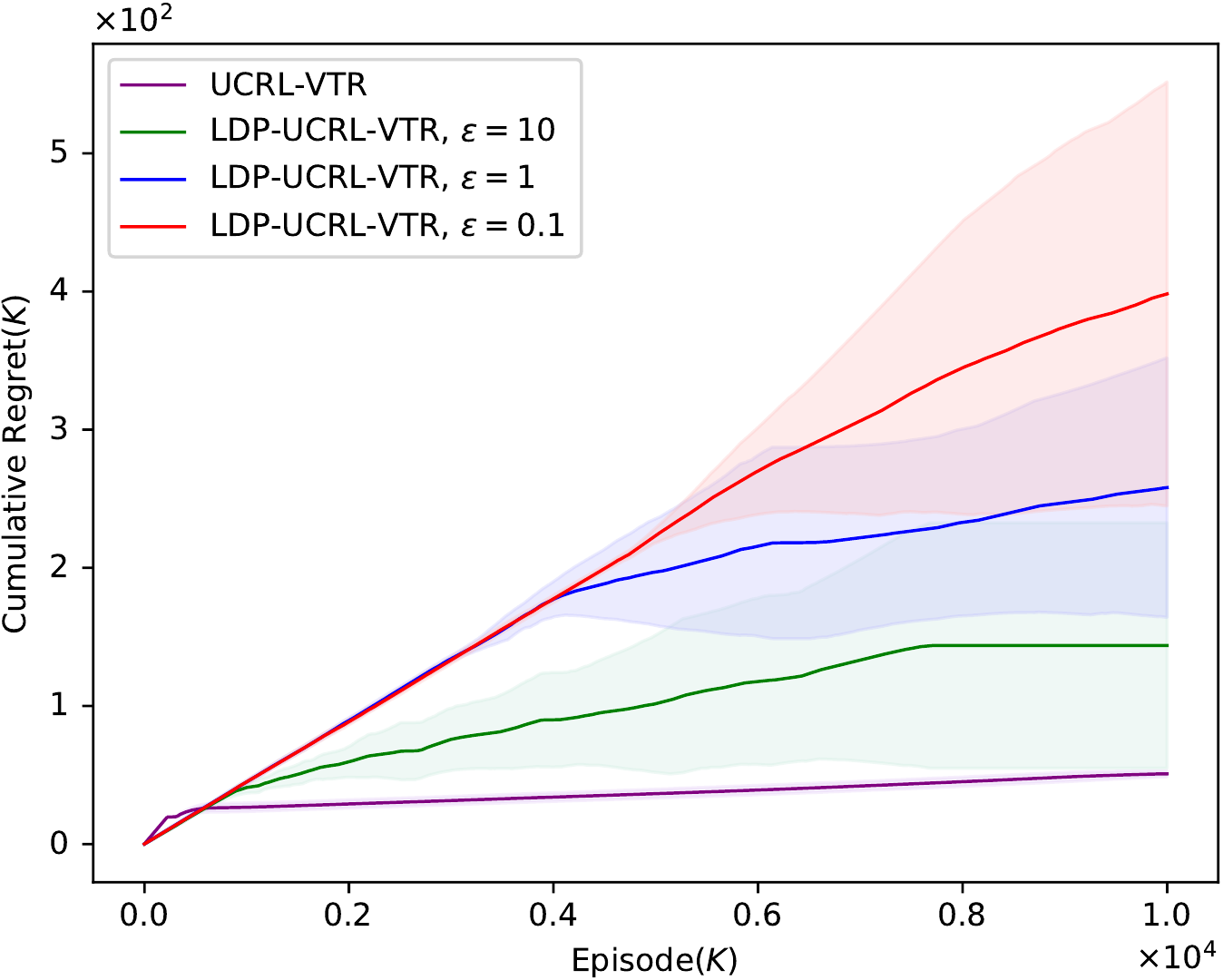}
    }\
    \subfigure[$S=3, A=2, H = 6, d=18$, inhomogeneous]{%
      \includegraphics[width=0.44\linewidth]{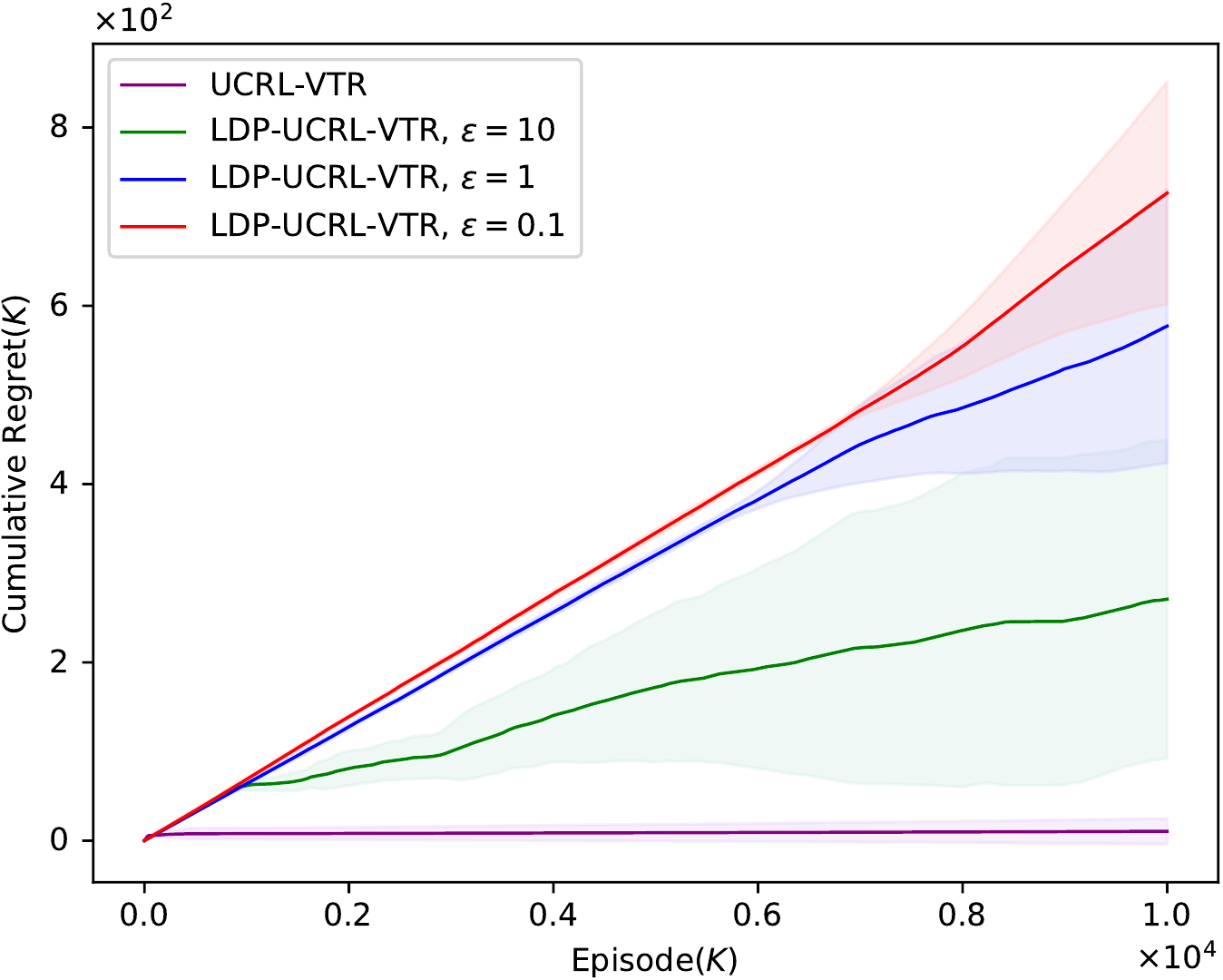}
    }
    \hfill
    \subfigure[$S=5, A = 2, H = 10, d=50$, inhomogeneous]{%
      \includegraphics[width=0.44\linewidth]{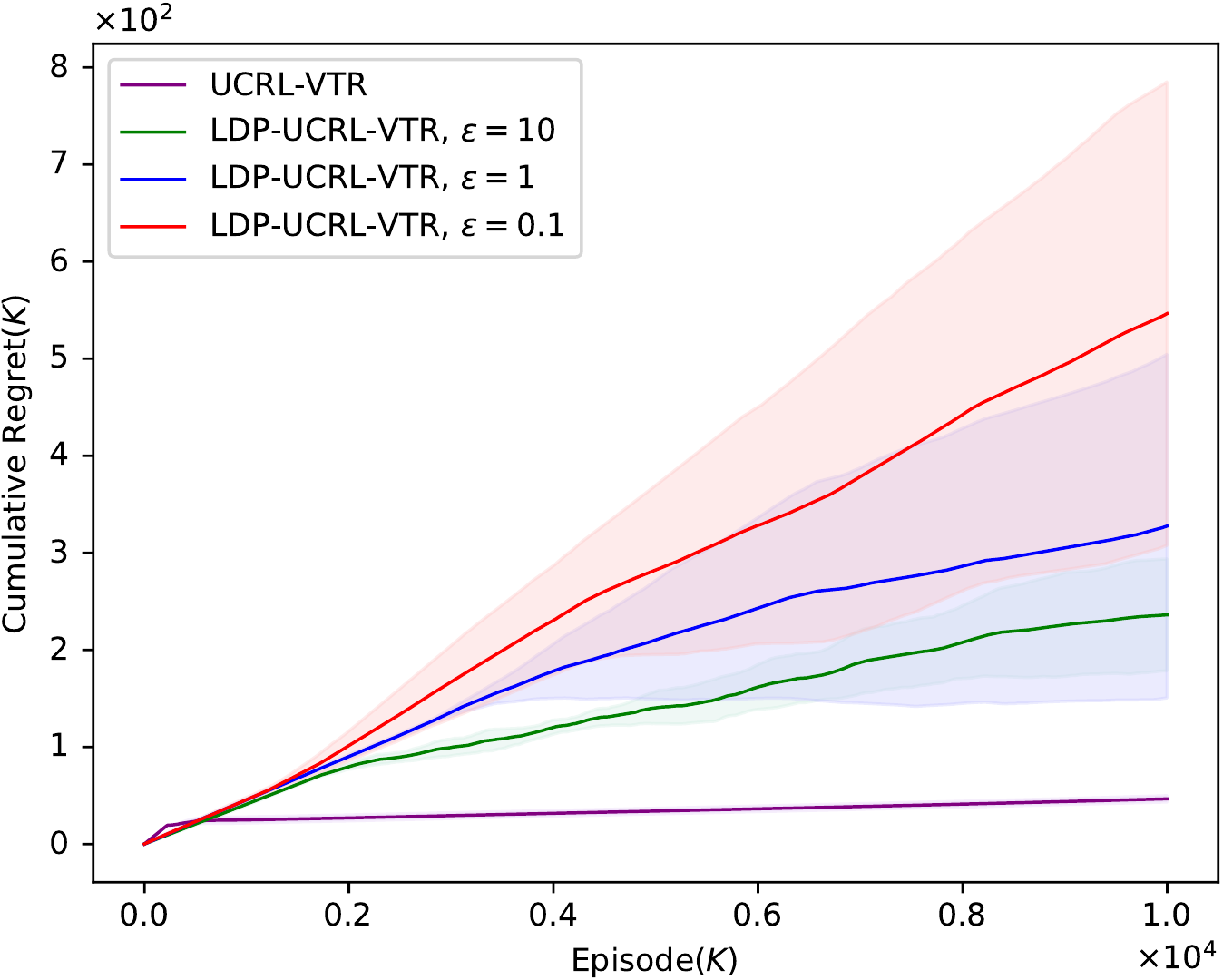}
    }
    \caption{Evaluation of the algorithms in two ``RiverSwim'' MDPs. Results are averaged over $10$ runs and the standard deviations are calculated to plot the confidence band. These results show that the cumulative regret of LDP-UCRL-VTR is sublinear in $K$, and its performance is getting closer to that of UCRL-VTR while the privacy guarantee becomes weaker, i.e., choosing a larger $\varepsilon$.}
    \label{fig:result}
\end{figure*}

\section{Experiments}
In this section, we carry out experiments to evaluate the performance of LDP-UCRL-VTR, and compare with its non-private counterpart UCRL-VTR \citep{jia2020model}.

\subsection{Experimental Setting} 
% \qg{Add back the figure of MDP construction}
We tested LDP-UCRL-VTR on a benchmark MDP instance named ``RiverSwim'' \citep{strehl2008analysis,ayoub2020model}. The purpose of this MDP is to tempt the agent to go left while it is hard for a short sighted agent to go right since $r(0, 0) \ne 0, r(s, 1) = 0, 0 \le s \le \abr{\cS}-1$. Therefore, it is hard for the agent to decide which direction to choose. In our experiment, the reward in each stage is normalized by $H$, e.g., $r(0, 0) = 5/(1000H), r(S, 1) = 1/H$. Our LDP-UCRL-VTR is also tested on the time-inhomogeneous ``RiverSwim'', where for each $h \in [H]$, the transition probability $p_h$ is sampled from a uniform distribution $U(0.8, 1)$. We also 
choose $H = 2S$. Figure \ref{fig:mdp} shows the state transition graph of this MDP.
% \tikzset{global scale/.style={
%     scale=#1,
%     every node/.append style={scale=#1}
%   }
% }
\subsection{Results and Discussion}
We evaluate LDP-UCRL-VTR with different privacy budget $\varepsilon$ and compare it with UCRL-VTR on both homogeneous and inhomogeneous ``RiverSwim''. For UCRL-VTR, we set $\sqrt{\beta} = c\sqrt{d_1} + (H - h + 1) \sqrt{2\log(1/K)+\log \det \Mb}$, where $d_1 = SA$ with $S$ being the number of states and $A$ being the number of actions, $\Mb$ is the covariance matrix in Algorithm 3 \citep{jia2020model}. We fine tune the hyper parameter $c$ for different experiments. For LDP-UCRL-VTR, we choose $\delta = 0.1, \alpha = 0.01$. Since $\delta$ and $\alpha$ are prefixed for all experiments, they can be treated as constants. Thus, we can choose $\beta$ in the form $cd^{3/4}(H - h + 1)^{3/2}k^{1/4}$ and only fine tune $c$. The results for each $\varepsilon$ are averaged over $10$ runs.

In our experiments, since the reward is normalized by $H$, we need to recompute $\sigma$ for the Gaussian mechanism. Recall that $\sigma = 2\Delta f H\sqrt{2\log(2.5H/\delta)}/\varepsilon$, where $\Delta f$ represents the $\ell_2$ sensitivity of $\Delta\ub$ in Algorithm \ref{algorithm}. In our setting, $\abr{Q}\le 1$ and therefore $\Delta f\le 1$. Thus, we set $\sigma = 4 H\sqrt{2\log(2.5H/\delta)}/\varepsilon$. In addition, we set $K = 10000$ for all experiments. To fine tune the hyper parameter $c$, we use grid search and select the one which attains the best result. The experiment results are shown in Figure \ref{fig:result}.

From Figure \ref{fig:result}, we can see that the cumulative regret of LDP-UCRL-VTR is indeed subliear in $K$. In addition, it is not surprising to see that LDP-UCRL-VTR incurs a larger regret than UCRL-VTR. The performance of LDP-UCRL-VTR with larger $\varepsilon$ is closer to that of UCRL-VTR as the privacy guarantee becomes weaker. Our results are also greatly impacted by $H$ and $d$, as the convergence (learning speed) slows down as we choose larger $H$ and smaller $\varepsilon$. The experiments are consistent with our theoretical results.

\section{Conclusion and Future Work}\label{sectionDiscussion}
In this paper, we studied RL with linear function approximation and LDP guarantee. To the best of our knowledge, our designed algorithm is the first provable privacy-preserving RL algorithm with linear function approximation. We proved that LDP-UCRL-VTR satisfies $(\varepsilon, \delta)$-LDP. We also show that LDP-UCRL-VTR enjoys an $\tilde{\cO}( d^{5/4}H^{7/4}T^{3/4}\rbr{\log(1/\delta)}^{1/4}\sqrt{1/\varepsilon})$ regret. Besides, we proved a lower bound $\Omega(dH\sqrt{T}/\rbr{e^{\varepsilon}(e^{\varepsilon}-1)})$ for $\varepsilon$-LDP linear mixture MDPs. We also provide experiments on synthetic datasets to corroborate our theoretical findings. 

In our current results, there is still a gap between the regret upper bound and the lower bound. We conjecture the gap to be a fundamental difference between learning linear mixture MDPs and tabular MDPs. In the future, it remains to study if this gap could be eliminated. 

%%%%%%%%%
% \newpage
\appendix

\section{Proof of the Privacy Guarantee}\label{appendLDP}
In this section, we provide the proof of Theorem \ref{LDPtheorem} with the help of the Gaussian mechanism, which is introduced in Lemma \ref{gaussian mechanism}. We start with the following definition.
\begin{definition}
    (Privacy Loss, \citealt{dwork2014algorithmic, abadi2016deep}). For neighboring databases $d,d^\prime \in \cD^n$, a mechanism $\cM$, auxiliary input $\textbf{aux}$, and an outcome $o\in \cR$, we define the privacy loss at outcome $o$ as 
    \[c(o; \cM, \textbf{aux}, d, d^\prime):= \log \frac{\PP\sbr{\cM(\textbf{aux}, d)=o}}{\PP\sbr{\cM(\textbf{aux}, d^\prime)=o}}\]
\end{definition}
With the definition of Privacy Loss $c$, the following theorem can provide a guarantee of the $(\varepsilon, \delta)$-DP property.
\begin{theorem}\label{DP}
    If the privacy loss $c$ satisfies that $\PP_{o \sim \cM(d)}\sbr{c(o; \cM, \textbf{aux}, d, d^\prime)>\varepsilon}\le \delta$ for all auxiliary input $\textbf{aux}$ and neighboring databases $d,d^\prime \in \cD^n$, then the mechanism $\cM$ satisfies $(\varepsilon, \delta)$-DP property.
\end{theorem}

\begin{proof}[Proof of Theorem \ref{DP}]
    This proof share the similar structure as that in \citet{abadi2016deep}. For simplicity, we denote the set $\cB$ as
    \[\cB = \cbr{o:c(o; \cM, \textbf{aux}, d, d^\prime)>\varepsilon}\,,\]
    which contain all ``bad'' outcome. With this notation, for each set $S$, we have
     \begin{align*}
    \PP\sbr{\cM(D)\in S} &= \PP\sbr{\cM(D)\in(S \cap \cB)}+ \PP\sbr{\cM(D)\in(S \setminus \cB)}\\
    &\leq \PP\sbr{\cM(D)\in \cB}+ \PP\sbr{\cM(D)\in(S \setminus \cB)}\\
    & \leq \PP\sbr{\cM(D)\in \cB}+ e^\varepsilon\PP\sbr{\cM(D^\prime)\in(S \setminus \cB)}\\
    & \leq \PP\sbr{\cM(D)\in \cB}+ e^\varepsilon\PP\sbr{\cM(D^\prime)\in S}\,,
    \end{align*}
    where the first inequality holds due to the monotone property of probability measure with $(S\cap\cB) \subseteq \cB$, the second inequality holds due to the definition of set $\cB$ and the last inequality holds due to the monotone property of probability measure with $( S \setminus \cB ) \subseteq S$. In addition, since $\PP_{o \sim \cM(d)}\sbr{c(o; \cM, \textbf{aux}, d, d^\prime)>\varepsilon}\le \delta$, we have
    \[\PP\sbr{\cM(D)\in S}\le \delta + e^\varepsilon\PP\sbr{\cM(D^\prime)\in S}\,,\]
and it implies that the mechanism $\cM$ satisfies $(\varepsilon, \delta)$-DP property.
\end{proof}
Now we prove Theorem \ref{LDPtheorem}.

\theoremstyle{remark}
\newtheorem*{LDPproof}{Proof of Theorem \ref{LDPtheorem}}
\begin{LDPproof} To prove the Theorem \ref{LDPtheorem}, 
it suffices to prove that for each episode $k\in {K}$, Algorithm~\ref{algorithm} satisfies the $(\varepsilon,\delta)$-LDP property. In the following proof, to avoid cluttered notation, we omit the superscript $k$ for simplicity. For the Gaussian mechanism, we first compute the sensitivity coefficient $\ell_2$ for Algorithm \ref{algorithm}. We denote $\Delta\tilde{\ub}_h= \bphi_{V_{k, h+1}}(s_h^k,a_h^k)V_{k, h+1}(s_{h+1}^k)$ and $\Delta\tilde{\bLambda}_h= \bphi_{V_{k, h+1}}(s_h^k,a_h^k)\bphi_{V_{k, h+1}}(s_h^k,a_h^k)^{\top}$, where $V_{k, h+1}(s_{h+1}^k)$ is a scalar. Therefore, for the vector $\Delta\tilde{\ub}_h$, the sensitivity coefficient is upper bound by
    \begin{align*}
        \nbr{\Delta\tilde{\ub}_h-(\Delta\tilde{\ub}_h)^\prime}_{2}  \leq \nbr{\bphi_{V_{k, h+1}}}_{2}\left\vert V_{k, h+1} \right\vert + \nbr{\bphi_{V_{k, h+1}}^\prime}_{2} \left\vert V_{k, h+1}^\prime \right\vert 
        \leq 2H^2\,,
    \end{align*}
    where the first inequality holds due to that fact that $\|\xb+\yb\|_2\leq \|\xb\|_2 + \|\yb\|_2$ and the last inequality holds due to \eqref{linearMixure}. Similar, for the matrix $\Delta\tilde{\bLambda}_h= \bphi_{V}(s_h^k,a_h^k)\bphi_{V}(s_h^k,a_h^k)^{\top}$, the sensitivity coefficient is upper bound by
        \begin{align*}
        \nbr{\bphi_V\bphi_V^\top - \bphi_V^{\prime}\bphi_V^{\prime\top}}_F &\le \nbr{\bphi_V\bphi_V^{\top}}_F+\nbr{\bphi_V^{\prime}\bphi_V^{\prime\top}}_F\\
        &=\sqrt{\tr\sbr{\bphi_V\bphi_v^{\top}\bphi_V\bphi_V^{\top}}} + \sqrt{\tr\sbr{\bphi_V^{\prime}\bphi_V^{\prime\top}\bphi_V^{\prime}\bphi_V^{\prime\top}}}\\
        &=\bphi_V^{\top}\bphi_V+\bphi_V^{\prime\top}\bphi_V^{\prime}\\
        &\le 2H^2\,,
    \end{align*}
    where the first inequality holds due to that fact that $\|\xb+\yb\|_2\leq \|\xb\|_2 + \|\yb\|_2$ and the last inequality holds due to \eqref{linearMixure}.
    According to the Algorithm \ref{algorithm} (Lines \ref{11} and \ref{12}), we have 
$\Delta \bLambda _{h}=\Delta \tilde{\bLambda}_h+\Wb_h$ and $ \Delta \ub _{h}=\Delta \tilde{\ub}_h+\bxi_h$, where $\Wb_h$ are independent symmetric Gaussian matrices  and $\bxi_h$ are independent Gaussian vector defined in the Algorithm 1. Now, We use $D_{1:h}$ to denote the collected information from stage $1$ to stage $h$. Considering two different datasets $D_{h}$, $(D_{h})^{\prime}$ collected by the server and any possible outcome $(\Mb, \balpha)$ of the Algorithm \ref{algorithm}, then we have
    \begin{align*}
        &\frac{\PP\rbr{\forall h \in [H], \rbr{\Delta \bLambda _{h},\Delta \ub _{h}}=\rbr{\Mb, \balpha}\given \cbr{D_h}_{h=1}^H}}{\PP\rbr{\forall h \in [H], \rbr{(\Delta \bLambda _{h})^\prime,(\Delta \ub _{h})^\prime}=\rbr{\Mb, \balpha}\given \cbr{D_h^\prime}_{h=1}^H}}\\
        &=\prod_{h=1}^H\frac{\PP\rbr{\rbr{\Wb_h^k,\bxi_h^k}=\rbr{\Mb-\Delta \tilde{\bLambda}_h, \balpha-\Delta \tilde{\ub}_{h}}\given D_{1:h-1}}}{\PP\rbr{\rbr{(\Wb_h)^\prime,(\bxi_h)^\prime}=\rbr{\Mb-(\Delta \tilde{\bLambda}_h)^\prime, \balpha-(\Delta \tilde{\ub}_h)^\prime}\given D^{\prime}_{1:h-1}}}\,,
    \end{align*}
    where the equation holds due to Markov property. With the help of the Markov property, we can further decompose the probability as
    \begin{align*}
        &\prod_{h=1}^{H} \frac{\PP\rbr{(\Wb_h,\bxi_h)=(\Mb-\Delta \tilde{\bLambda}_h, \balpha-\Delta \tilde{\ub}_h)\given D_{h-1}}}{\PP\rbr{\rbr{(\Wb_h)^\prime,(\bxi_h)^\prime}=\rbr{\Mb-(\Delta \tilde{\bLambda}_h)^\prime, \balpha-(\Delta \tilde{\ub}_h)^\prime}\given D_{h-1}^\prime}} \\
        &=\prod_{h=1}^{H}\frac{\PP\rbr{\Wb_h=\Mb-\Delta \tilde{\bLambda}_h\given D_{h-1}}\PP\rbr{\bxi_h=\balpha-\Delta \tilde{\ub}_h \given D_{h-1}}}{\PP\rbr{(\Wb_h)^\prime=\Mb-(\Delta \tilde{\bLambda}_h)^\prime\given D_{h-1}^\prime}
        \PP\rbr{(\bxi_h)^\prime)=\balpha-(\Delta \tilde{\ub}_h)^\prime\given D_{h-1}^\prime}}\,.
    \end{align*}
According to Lemma \ref{gaussian mechanism} and the sensitivity coefficient of  $\Delta \tilde{\bLambda}_h$, if we set $\sigma=4H^3\sqrt{2\log(2.5H/\delta)}/\varepsilon$, then  with probability at least $1-\delta/(2H)$, for the $\Wb_h$ term, we have
    \[\PP\rbr{\Wb_h=\Mb-\bphi_V\bphi_V^\top\given D_{h-1}}\le \exp\rbr{\frac{\varepsilon}{2H}} \PP\rbr{(\Wb_h)^\prime=\Mb-\bphi_V^\prime\bphi_V^{\prime\top}\given D_{h-1}^\prime}\,.\]
    For the $\bxi$ term,  the probability density function (PDF) can be written as
    \[\frac{\PP\rbr{\bxi_h=\balpha-\Delta \tilde{\ub}_h \given D_{h-1}}}{\PP\rbr{(\bxi_h)^\prime=\balpha-(\Delta \tilde{\ub}_h)^\prime\given D_{h-1}^\prime}}=\frac{\exp \rbr{-\nbr{\balpha-\Delta \tilde{\ub}_h}^2/(2\sigma^2)}}{\exp \rbr{-\nbr{\balpha-\Delta \tilde{\ub}_h + \rbr{\Delta \tilde{\ub}_h - (\Delta \tilde{\ub}_h)^\prime}}^2/(2\sigma^2)} }\,,\]
Therefore, applying Lemma \ref{gaussian mechanism}, with probability at least $1-\delta/(2H)$, we have the following inequality
    \[\PP\rbr{\bxi_h^k=\balpha-\Delta \tilde{\ub}_h \given \cbr{D_h}_{h=1}^H}\le \exp\rbr{\frac{\varepsilon}{2H}}\PP\rbr{(\bxi_h^k)^\prime=\balpha-(\Delta \tilde{\ub}_h)^\prime\given \cbr{D_h^\prime}_{h=1}^H}\,.\]
Finally, taking a union bound for $\Wb_h$,$\bxi$ terms and all stage $h\in[H]$, with probability at least $1-2H\times \delta/(2H)=1-\delta$, we have 
    \[\log\sbr{\frac{\PP\rbr{\forall h \in [H], \rbr{\Delta \bLambda _{h},\Delta \ub _{h}}=\rbr{\Mb, \balpha}\given \cbr{D_h}_{h=1}^H}}{\PP\rbr{\forall h \in [H], \rbr{(\Delta \bLambda _{h})^\prime,(\Delta \ub _{h})^\prime}=\rbr{\Mb, \balpha}\given \cbr{D_h^\prime}_{h=1}^H}}}\le \varepsilon\,.\]
    Therefore, according to Theorem \ref{DP}, we can conclude that our algorithm protects $(\varepsilon, \delta)$-LDP property. According to the post-processing property, one can also prove that our algorithm satisfies DP property.
\end{LDPproof}
\section{Proofs of Regret Upper Bound}\label{appenMainTheorem}
In this section, we provide the proof of Theorem \ref{mainTheorem}. We first propose the following lemmas.
\begin{lemma}\label{main}
If we choose parameter $\lambda=1$ and $\beta=c^\prime d^{3/4}H^{3/2}k^{1/4}\log(dT/\alpha)\big(\log(H/\delta)\big)^{1/4}\sqrt{1/\varepsilon}$ for a large enough constant $c^\prime$ in Algorithm \ref{algorithm}, then for any fixed policy $\pi$ and all pairs $(s, a, h, k) \in \cS \times \cA \times[H]\times[K]$, with probability at least $1-\alpha/2$, we have
    \[\nbr{(\bSigma_{k, h})^{1/2}\rbr{\hat{\btheta}_{k,h}-\btheta_{h}^{*}}}\le \beta\,.\]
\end{lemma}
\begin{proof}[Proof of Lemma \ref{main}]
According to the definition of $\hat{\btheta}_{k,h}$ in Algorithm \ref{algorithm} (Line \ref{19}), the difference between our estimator $\hat{\btheta}_{k,h}$ and underlying vector $\btheta_{h}^{*}$ can be decomposed as
    \begin{align}
        \hat{\btheta}_{k,h}-\btheta_{h}^{*}&=(\bSigma_{k, h})^{-1}\sum_{\tau=1}^{k-1} \cbr{\bphi_{V_{\tau, h+1}}(s_h^\tau, a_h^\tau)V_{\tau, h + 1}(s_{h+1}^{\tau})+\bxi_h^{\tau}}-\btheta_{h}^{*}\notag\\
        &= (\bSigma_{k, h})^{-1}\left\{ -\lambda \btheta_{h}^{*}-\sum_{\tau=1}^{k-1}\bphi_{V_{\tau, h+1}}\bphi_{V_{\tau, h+1}}^\top \btheta_{h}^{*} - \Wb^{k-1} \btheta_{h}^{*}+\sum_{\tau=1}^{k-1}\bphi_{V_{\tau, h+1}}V_{\tau, h + 1}+\sum_{\tau=1}^{k-1}\bxi_h^{\tau}\right\}\notag \\
        &= (\bSigma_{k, h})^{-1}\cbr{\underbrace{\rbr{-\lambda \mathbf{I}-\Wb^{k-1}}\btheta_{h}^{*}}_{\qb_1}+\underbrace{\sum_{\tau=1}^{k-1} \bphi_{V_{\tau, h+1}}\sbr{V_{\tau, h + 1} - \PP_h V_{\tau,h+1}}}_{\qb_2}+\underbrace{\sum_{\tau=1}^{k-1}\bxi_h^{\tau}}_{\qb_3}}\,,\notag
    \end{align}
    where the matrix $\Wb^{k-1}=\sum_{i=1}^{k-1}\Wb_{i,h}$. Simply rewriting the above equation, we have
    \begin{align}
             \nbr{(\bSigma_{k, h})^{1/2}\rbr{\hat{\btheta}_{k,h}-\btheta_{h}^{*}}} = \nbr{(\bSigma_{k, h})^{-1/2}(\qb_1+\qb_2+\qb_3)}\,. \label{eq:000}   
    \end{align}
    Now, we first give a upper and lower bounds for the eigenvalues of the symmetric Gaussian matrix
    $\Wb^{k-1}$. According to the Algorithm \ref{algorithm}  (Line \ref{11}), all entries of $\Wb_{i,h}$ are sampled from $\mathcal{N}(0, \sigma^2)$ and known concentration results \citep{tao2012topics} on the top singular value shows that
    \[\PP\rbr{\nbr{\sum_{\tau = 1}^{k-1}\Wb_h}\ge \sqrt{k-1}\sigma\rbr{\sqrt{4d}+2\log(6H/\alpha)}}\le \frac{\alpha}{6H}\,.\]
    For simplicity, we denote $\Gamma = \sqrt{k-1}\sigma(\sqrt{4d}+2\log(6H/\alpha))$. Since the symmetric Gaussian matrix $\Wb^{k-1}$ may not preserve PSD property, we use the shifted regularizer. More specifically, after adding a basic matrix $2\Gamma\Ib$ to the matrix $\Wb^{k-1}=\sum_{i = 1}^{k-1}\Wb_{i,h}$, for each stage $h \in [H]$, the eigenvalues can be bounded by the interval $[\Gamma, 3\Gamma]$ with probability at least $1-\alpha/6$. We use the notation $\cE_1$ to denote the event that
    \[\forall h \in [H], \qquad \forall j \in [d], \Gamma\le\sigma_{j}\le 3\Gamma\,,\]
    where $\sigma_1,..,\sigma_d$ are eigenvalue of the matrix $\Wb^{k-1}$
    and we have  $\PP(\cE_1) \ge 1-\alpha/6$.
    
    Now, we assume the event $\cE_1$ holds and we further denote $\rho_{\max} = 3\Gamma + \lambda, \rho_{\min} = \Gamma + \lambda$. Then for the term $q_1$, we have
    \begin{align}
        \nbr{(\bSigma_{k, h})^{-1/2} \qb_1} & \leq \nbr{(\Wb^{k-1} + \lambda \Ib)^{-1/2} \qb_1}\notag\\
        & = \nbr{(\Wb^{k-1} + \lambda \Ib)^{1/2}\btheta_{h}^{*}}\notag\\
        & \leq \sqrt{\rho_{\max}} \nbr{\btheta_{h}^{*}}\notag\\
        & \le \sqrt{12d\sqrt{k-1}H^3\sqrt{2\log(2.5H/\delta)}(4\sqrt{d}+2\log(6H/d))/\varepsilon+\lambda d}\,,\label{eq:001}
    \end{align}
    where the first inequality holds due to $\bSigma_{k, h}\succeq \Wb^{k-1} + \lambda \Ib$, the second inequality holds due to the definition of event $\cE_1$ and the last inequality holds due to the choice of $\sigma$ in Algorithm (Line \ref{1}).
    
    For the term $q_2$, we have
    \[\nbr{(\bSigma_{k, h})^{-1/2}\qb_2} = \nbr{\sum_{\tau=1}^{k-1} \bphi_{V_{\tau, h+1}}\sbr{V_{\tau, h + 1} - \PP_h V_{\tau,h+1}}}_{(\bSigma_{k, h})^{-1}}\le \nbr{\sum_{\tau=1}^{k-1} \bphi_{V_{\tau, h+1}}\sbr{V_{\tau, h + 1} - \PP_h V_{\tau,h+1}}}_{(\Zb)^{-1}}\,, \]
    where $\Zb = \lambda\Ib + \sum_{\tau=1}^{k-1}\bphi_{V_{\tau, h+1}}\bphi_{V_{\tau, h+1}}^\top$ and the inequality holds due to the tact that  $\bSigma_{k, h}\succeq \Zb = \lambda\Ib + \sum_{\tau=1}^{k-1}\bphi_{V_{\tau, h+1}}\bphi_{V_{\tau, h+1}}^\top$. According to the Definition \ref{definMDP}, we have $V_{\tau,h + 1} = \PP_h V_{\tau,h + 1} + \eta_{\tau, h + 1} = \bphi_{V_{\tau, h + 1}}^\top\btheta^*_h + \eta_{\tau, h + 1} $. Let $\cbr{\cG_t}_{t=1}^\infty$ be a filtration, $\cbr{\bphi_{V_{\tau, t}}, \eta_{\tau, t}}_{t \ge 1}$ a stochastic process so that $\bphi_{V_{\tau, t}}$ is $\cG_t$-measurable and $\eta_{\tau, t}$ is $\cG_{t + 1}$-measurable. With this notation, we further have
    \[\abr{\eta_t}\le |V_{\tau, t}-\PP_h V_{\tau,h + 1}|\le H, \qquad \EE[\eta_t^2\given \cG_t] \le \EE[V_{\tau, t}^2\given \cG_t] \le H^2\,.\]
    Now  we introduce the following event $\cE_2$
    \[\cE_2 = \cbr{\forall h\in[H], \nbr{\qb_2}_{\Zb^{-1}}\le 4H\rbr{2\sqrt{d\log\rbr{1+\frac{(k-1)H^2}{d\lambda}}\log\rbr{\frac{24(k-1)^2H}{\alpha}}} + \log\rbr{\frac{24(k-1)^2H}{\alpha}}}}\,,\]
    then from Theorem 4.1 in \citet{zhou2020nearly}, we have $\cE_2$ holds with probability $1-\alpha/6$.
    
    For the  term $q_3$, it can be upper bounded by 
    \begin{align*}
        \nbr{\sum_{\tau=1}^{k-1}\bxi_h^{\tau}}_{(\bSigma_{k, h})^{-1}}&\le \nbr{\sum_{\tau=1}^{k-1}\bxi_h^{\tau}}_{\rbr{\Wb^{k-1} + \lambda \Ib}^{-1}}\le\frac{1}{\sqrt{\rho_{\min}}}
        \nbr{\sum_{\tau=1}^{k-1}\bxi_h^{\tau}}_2\,,
    \end{align*}
    where the first inequality holds due to $\bSigma_{k, h}\succeq \Wb^{k-1} + \lambda \Ib$ and the second inequality holds due to the definition of event $\cE_1$. Furthermore, by Lemma \ref{hoeffding}, with probability at least $1-\alpha/6H$, there is
    \[\nbr{\sum_{\tau=1}^{k-1}\bxi_h^{\tau}}_{2} \leq \sigma\sqrt{(k-1)d\log\frac{12dH}{\alpha}}\,.\]
    We also let $\cE_3$ be the event that:
\[\cE_3= \cbr{\forall h \in [H]:\qquad \nbr{\sum_{\tau=1}^{k-1}\bxi_h^{\tau}}_{2} \leq \sigma\sqrt{(k-1)d\log\frac{12dH}{\alpha}}}\,.\]

    By taking a union bound for all stage $h\in[H]$, we have $\PP(\cE_3)\ge 1-\alpha/6$. Therefore,
    \begin{align}
        \nbr{\sum_{\tau=1}^{k-1}\bxi_h^{\tau}}_{(\bSigma_{k, h})^{-1}}&\le\frac{1}{\sqrt{\rho_{\min}}}
        \nbr{\sum_{\tau=1}^{k-1}\bxi_h^{\tau}}_2\notag\\
        &\le \frac{\sigma\sqrt{(k-1)d\log(12dH/\alpha)}}{\sqrt{\rho_{\min}}}\notag\\
        &= \frac{\sigma\sqrt{(k-1)d\log(12dH/\alpha)}}{\sqrt{\sigma\sqrt{(k-1)}(\sqrt{4d}+2\log(6H/\alpha)) + \lambda}}\notag\\
        &\le \frac{\sqrt{\sigma}(k-1)^{1/4}\sqrt{d\log(12dH/\alpha)}}{\sqrt{\sqrt{4d}+2\log(6H/\alpha)}}\notag\\
        &\le d^{1/4}(k-1)^{1/4}H^{3/2}\sqrt{2\sqrt{2\log(2.5H/\delta)}/\varepsilon}\sqrt{\log(12dH/\alpha)}\,,\label{eq:002}
    \end{align}
    where the second inequality holds due to the definition of event $\cE_3$, the third inequality holds due to $\lambda \ge 0$ and the last inequality holds by the fact that $2\log(6H/\alpha)\ge 0$. Finally, substituting the \eqref{eq:001}, \eqref{eq:002} with the definition of event $\cE_2$ into \eqref{eq:000},  with probability at least $1-\alpha/2$, we have for each $h \in [H]$, 
    \[\nbr{(\bSigma_{k, h})^{1/2}\rbr{\hat{\btheta}_{k,h}-\btheta_{h}^{*}}}\le \beta\,,\]
where $\beta=C^\prime d^{3/4}H^{3/2}k^{1/4}\log(dT/\alpha)\big(\log(H/\delta)\big)^{1/4}\sqrt{1/\varepsilon}$
and $C^\prime$ is an absolute constant.
\end{proof}

% \begin{lemma}
%     (Lemma B.5, \citealt{jin2020provably}). On the event defined in Lemma \ref{p2}, with probablity $1-\alpha$, we have $Q_h^k(s,a)\ge Q_h^*(s,a)$ for all $(s,a,h,k) \in \cS\times \cA\times [H]\times [K].$
% \end{lemma}
\begin{lemma}\label{optimistic}
    Let $Q_{k,h}, V_{k,h}$ be the value functions defined in Algorithm \ref{algorithm}. Then, on the event $\cE_1, \cE_2, \cE_3$, defined in Lemma \ref{main}, for any pairs $(s, a, k, h) \in \cS \times \cA \times [K] \times [H]$, we have  $Q^*_h(s, a) \le Q_{k,h}(s, a)$ and $ V_h^*(s) \le V_{k,h}(s)$.
\end{lemma}
\begin{proof}[Proof of Lemma \ref{optimistic}]
We prove this lemma by induction. First, we consider for that basic case. The statement holds for $H + 1$ since $Q_{k,H+1}(\cdot, \cdot) = 0 = Q^*_{H+1}(\cdot, \cdot)$ and $V_{k,H+1}(\cdot) = 0 = V^*_{H+1}(\cdot)$. Now, suppose that this statement also holds for $h + 1$, we have $Q_{k, h+1}(\cdot, \cdot) \geq Q_{h+1}^{*}(\cdot, \cdot), V_{k, h+1}(\cdot) \geq V_{h+1}^{*}(\cdot)$. For stage $h$ and state-action pair $(s,a)$, if $Q_{k,h}(s,a)\ge H$, obviously, we have $Q_{k,h}(s,a)\ge H\ge Q_h^*(s,a)$. Otherwise, we have
\begin{align*}
Q_{k, h}(s, a)-Q_{h}^{*}(s, a) 
&= r_h(s,a) + \left<\widehat{\boldsymbol{\theta}}_{k, h}, \boldsymbol{\phi}_{V_{k, h+1}}\left(s, a\right)\right> + {\beta}_{k}\left\|{\boldsymbol{\Sigma}}_{k, h}^{-1 / 2} \boldsymbol{\phi}_{V_{k, h+1}}\left(s, a\right)\right\|_{2}\\
&\qquad- r_h(s, a)-\left[\mathbb{P}_{h} V_{h+1}^*\right]\left(s, a\right)\\
&= \left<\widehat{\boldsymbol{\theta}}_{k, h}, \boldsymbol{\phi}_{V_{k, h+1}}\left(s, a\right)\right> - \left<\btheta^*_h, \boldsymbol{\phi}_{V_{k, h+1}}\left(s, a\right)\right> + \left<\btheta^*_h, \boldsymbol{\phi}_{V_{k, h+1}}\left(s, a\right)\right>\\
&\qquad  + {\beta}_{k}\left\|{\boldsymbol{\Sigma}}_{k, h}^{-1 / 2} \boldsymbol{\phi}_{V_{k, h+1}}\left(s, a\right)\right\|_{2} - \left[\mathbb{P}_{h} V_{h+1}^*\right]\left(s, a\right)\\
&=  {\beta}_{k}\left\|{\boldsymbol{\Sigma}}_{k, h}^{-1 / 2} \boldsymbol{\phi}_{V_{k, h+1}}\left(s, a\right)\right\|_{2} - \left<\btheta^*_h - \widehat{\boldsymbol{\theta}}_{k, h}, \boldsymbol{\phi}_{V_{k, h+1}}\left(s, a\right)\right>\\
&\qquad +\PP V_{k, h+1}(s,a) - \PP V_{h+1}^*(s,a)\\
&\ge {\beta}_{k}\left\|{\boldsymbol{\Sigma}}_{k, h}^{-1 / 2} \boldsymbol{\phi}_{V_{k, h+1}}\left(s, a\right)\right\|_{2} - \left\|{\boldsymbol{\Sigma}}_{k, h}^{1 / 2}\left(\hat{\btheta}_{k, h}-\boldsymbol{\theta}_{h}^{*}\right)\right\|_{2}\left\|{\boldsymbol{\Sigma}}_{k, h}^{-1 / 2} \boldsymbol{\phi}_{V_{k, h+1}}\left(s, a\right)\right\|_{2}\\
&\qquad +\PP V_{k, h+1}(s,a) - \PP V_{h+1}^*(s,a)\\
&\ge \PP V_{k, h+1}(s,a) - \PP V_{h+1}^*(s,a)\\
&\ge 0\,,
\end{align*}
where the first inequality is because of Cauchy-Schwarz inequality, the second inequality holds due to Lemma \ref{main}, and the last inequality holds by the induction assumption with the fact that $\PP_h$ is a
monotone operator with respect to the partial ordering of functions. Moreover, since we have $Q_{k, h}(\cdot, \cdot) \geq Q_{h}^{*}(\cdot, \cdot)$ for any state-action pair $(s,a)$, we directly obtains that $V_{k, h}(\cdot) \geq V_{h}^{*}(\cdot)$. Therefore, we conclude the proof of this lemma.
\end{proof}

Now we begin to prove our main Theorem.
\theoremstyle{remark}
\newtheorem*{mainUpper}{Proof of Theorem \ref{mainTheorem}}
\subsection{Proof of Theorem \ref{mainTheorem}}\label{proofOfMain}
\begin{mainUpper}
We give the proof of our main theorem on the events $\cE_1, \cE_2, \cE_3$ defined in Lemma \ref{main}. According to Lemma \ref{optimistic}, we have that $Q_h^*(s, a) \le Q_{k,h}(s,a), V_h^*(s)\le V_{k,h}(s)$. Thus, we have
\begin{align*}
V_{h}^*\rbr{s_{h}^{k}} - V_{h}^{\pi^{k}}\left(s_{h}^{k}\right)\le& V_{k, h}\left(s_{h}^{k}\right)-V_{h}^{\pi^{k}}\left(s_{h}^{k}\right)\\
=&\max_a Q_{k,h}(s_h^k, a) - \max_a Q_h^{\pi^k}(s_h^k,a)\\
\le&Q_{k,h}(s_h^k, a_h^k) - Q_h^{\pi^k}(s_h^k, a_h^k)\\
\leq& r_h(s_h^k,a_h^k) + \left<\widehat{\boldsymbol{\theta}}_{k, h}, \boldsymbol{\phi}_{V_{k, h+1}}\left(s_{h}^{k}, a_{h}^{k}\right)\right> + {\beta}_{k}\left\|{\boldsymbol{\Sigma}}_{k, h}^{-1 / 2} \boldsymbol{\phi}_{V_{k, h+1}}\left(s_{h}^{k}, a_{h}^{k}\right)\right\|_{2}\\
&\qquad- r_h(s_h^k, a_h^k)-\left[\mathbb{P}_{h} V_{h+1}^{\pi^{k}}\right]\left(s_{h}^{k}, a_{h}^{k}\right) \\
\leq &\left\|{\boldsymbol{\Sigma}}_{k, h}^{1 / 2}\left({\boldsymbol{\theta}}_{k, h}-\boldsymbol{\theta}_{h}^{*}\right)\right\|_{2}\left\|{\boldsymbol{\Sigma}}_{k, h}^{-1 / 2} \boldsymbol{\phi}_{V_{k, h+1}}\left(s_{h}^{k}, a_{h}^{k}\right)\right\|_{2} \\
&\qquad+\left[\mathbb{P}_{h} V_{k, h+1}\right]\left(s_{h}^{k}, a_{h}^{k}\right)-\left[\mathbb{P}_{h} V_{h+1}^{\pi^{k}}\right]\left(s_{h}^{k}, a_{h}^{k}\right)+{\beta}_{k}\left\|{\boldsymbol{\Sigma}}_{k, h}^{-1 / 2} \boldsymbol{\phi}_{V_{k, h+1}}\left(s_{h}^{k}, a_{h}^{k}\right)\right\|_{2} \\
\leq &\left[\mathbb{P}_{h} V_{k, h+1}\right]\left(s_{h}^{k}, a_{h}^{k}\right)-\left[\mathbb{P}_{h} V_{h+1}^{\pi^{k}}\right]\left(s_{h}^{k}, a_{h}^{k}\right)+2 {\beta}_{k}\left\|{\boldsymbol{\Sigma}}_{k, h}^{1 / 2} \boldsymbol{\phi}_{V_{k, h+1}}\left(s_{h}^{k}, a_{h}^{k}\right)\right\|_{2}\,,
\end{align*}
where the first inequality holds due to Lemma \ref{optimistic}, the second inequality holds due to the choice of action $a_h^k$ in the Algorithm (Line \ref{10}), the third inequality holds due to the definition of $V_{k,h}$ with the Bellman equation for $V_h^{\pi^k}$, the fourth inequality holds due to Cauchy-Schwartz inequality and the last inequality holds by Lemma \ref{main}.
We also note that $V_{k, h}\left(s_{h}^{k}\right)-V_{h}^{\pi^{k}}\left(s_{h}^{k}\right)\le V_{k, h}\left(s_{h}^{k}\right) \le H$ and it implies that
\begin{align*}
&V_{k, h}\left(s_{h}^{k}\right)-V_{h}^{\pi^{k}}\left(s_{h}^{k}\right)\\
&\leq \min \left\{H, 2 {\beta}_{k}\left\|{\boldsymbol{\Sigma}}_{k, h}^{1 / 2} \phi_{V_{k, h+1}}\left(s_{h}^{k}, a_{h}^{k}\right)\right\|_{2}+\left[\mathbb{P}_{h} V_{k, h+1}\right]\left(s_{h}^{k}, a_{h}^{k}\right)-\left[\mathbb{P}_{h} V_{h+1}^{\pi^{k}}\right]\left(s_{h}^{k}, a_{h}^{k}\right)\right\}\\
&\leq \min \left\{H, 2 {\beta}_{k}\left\|{\boldsymbol{\Sigma}}_{k, h}^{1 / 2} \boldsymbol{\phi}_{V_{k, h+1}}\left(s_{h}^{k}, a_{h}^{k}\right)\right\|_{2}\right\}+\left[\mathbb{P}_{h} V_{k, h+1}\right]\left(s_{h}^{k}, a_{h}^{k}\right)-\left[\mathbb{P}_{h} V_{h+1}^{\pi^{k}}\right]\left(s_{h}^{k}, a_{h}^{k}\right)\,,
\end{align*}
where the second inequality holds since $V_{k,h+1}\ge V_{h+1}^*\ge V_{h+1}^{\pi^k}$ and we further have
\begin{align*}
    &V_{k, h}\left(s_{h}^{k}\right)-V_{h}^{\pi^{k}}\left(s_{h}^{k}\right) - \sbr{V_{k, h + 1}\left(s_{h + 1}^{k}\right)-V_{h +1}^{\pi^{k}}\left(s_{h+1}^{k}\right)}\\
    &\le  \min \left\{H, 2 {\beta}_{k}\left\|{\boldsymbol{\Sigma}}_{k, h}^{1 / 2} \boldsymbol{\phi}_{V_{k, h+1}}\left(s_{h}^{k}, a_{h}^{k}\right)\right\|_{2}\right\}\\
    &\qquad+\left[\mathbb{P}_{h} V_{k, h+1}\right]\left(s_{h}^{k}, a_{h}^{k}\right)-\left[\mathbb{P}_{h} V_{h+1}^{\pi^{k}}\right]\left(s_{h}^{k}, a_{h}^{k}\right) - \sbr{V_{k, h + 1}\left(s_{h + 1}^{k}\right)-V_{h +1}^{\pi^{k}}\left(s_{h+1}^{k}\right)}\,.
\end{align*}
Summing up these inequalities for $k \in [K]$ and stage $h = h^\prime, \dots , H$
\begin{align*}
\sum_{k=1}^{K}\left[V_{k, h^{\prime}}\left(s_{ h^{\prime}}^k\right)-V_{h^{\prime}}^{\pi^{k}}\left(s_{ h^{\prime}}^k\right)\right] 
&\leq 2 \sum_{k=1}^{K} \sum_{h=h^{\prime}}^{H} {\beta}_{k} \min \left\{1,\left\|{\boldsymbol{\Sigma}}_{k, h}^{-1 / 2} \boldsymbol{\phi}_{V_{k, h+1}}\left(s_{h}^{k}, a_{h}^{k}\right) \right\|_{2}\right\} \\
&\quad+\sum_{k=1}^{K} \sum_{h=h^{\prime}}^{H}\left[\left[\mathbb{P}_{h}\left(V_{k, h+1}-V_{h+1}^{\pi^{k}}\right)\right]\left(s_{h}^{k}, a_{h}^{k}\right)-\left[V_{k, h+1}-V_{h+1}^{\pi^{k}}\right]\left(s_{h+1}^{k}\right)\right]\,.
\end{align*}
Now, we define the event $\cE_4$ as follows
\[\mathcal{E}_{4}=\left\{\forall h^{\prime} \in[H], \sum_{k=1}^{K} \sum_{h=h^{\prime}}^{H}\left[[\mathbb{P}_{h}(V_{k, h+1}-V_{h+1}^{\pi^{k}})]\left(s_{h}^{k}, a_{h}^{k}\right)-[V_{k, h+1}-V_{h+1}^{\pi^{k}}](s_{h+1}^{k})\right] \leq 4 H \sqrt{2 T \log (2H / \alpha)}\right\}\,.\]
Thus, since $[\mathbb{P}_{h}(V_{k, h+1}-V_{h+1}^{\pi^{k}})](s_{h}^{k}, a_{h}^{k})-[V_{k, h+1}-V_{h+1}^{\pi^{k}}](s_{h+1}^{k})$ forms a martingale difference sequence and $\abr{[\mathbb{P}_{h}(V_{k, h+1}-V_{h+1}^{\pi^{k}})](s_{h}^{k}, a_{h}^{k})-[V_{k, h+1}-V_{h+1}^{\pi^{k}}](s_{h+1}^{k})}\le 4H$, applying Azuma-Hoeffding inequality, we have $\cE_4$ holds with probability $\PP(\cE_4)\ge 1-\alpha/2$.

Now, note that $\bSigma \succeq \lambda \Ib$ and choosing the stage $h^{\prime} = 1$, by the Cauchy-Schwartz inequality, we have
\begin{align*}
    \sum_{k=1}^{K} \sum_{h=1}^{H} {\beta}_{k} \min \left\{1,\left\|{\boldsymbol{\Sigma}}_{k, h}^{-1 / 2} \boldsymbol{\phi}_{V_{k, h+1}}\left(s_{h}^{k}, a_{h}^{k}\right) \right\|_{2}\right\}&\le {\beta}_{K}\sum_{h=1}^{H}\sum_{k=1}^{K}\left\{1,\left\|{\boldsymbol{\Sigma}}_{k, h}^{-1 / 2} \boldsymbol{\phi}_{V_{k, h+1}}\left(s_{h}^{k}, a_{h}^{k}\right) \right\|_{2}\right\}\\
    &\le H \beta_K\sqrt{2dK\log(1 + K/\lambda)}\,,
\end{align*}
where the first inequality holds due to Cauchy-Schwartz and the last inequality holds due to Lemma~\ref{boundPhi}.

Finally, on the events $\cE_1, \cE_2, \cE_3, \cE_4$, we conclude with probability at least $1-\alpha$:
\begin{align*}
    \text{Regret}(K) &\le 4H\sqrt{2T\log(2H/\alpha)} + c d^{3/4}H^{3/2}K^{1/4}\log\frac{dT}{\alpha}\rbr{\log\frac{H}{\delta}}^{1/4}\sqrt{\frac{1}{\varepsilon}}H\sqrt{2dK\log(1 + K/\lambda)}\\
    &\le \tilde{\cO}\rbr{ d^{5/4}H^{5/2}K^{3/4}\rbr{\log\frac{1}{\delta}}^{1/4}\sqrt{\frac{1}{\varepsilon}}}\,.
\end{align*}
% where $c^{\prime}$ is an absolute constant.
\end{mainUpper}
\section{Proofs of Regret Lower Bound}\label{AppenC}

In this section, we provide the proof of lower bound for learning $\varepsilon$-LDP linear mixture MDPs, using the hard-to-learn MDP instance constructed in \citet{zhou2020nearly}. More specifically, there exist $H+2$ different states $s_1,..,s_{H+2}$, where $s_{H+1}$ and $s_{H+2}$ are absorbing states. The action space $\cA=\{-1,1\}^{d-1}$ consists of $2^{d-1}$ different actions. The reward function $r_h$ satisfies that $r_h(s_h,\ba)=0(1\leq h\leq H+1)$ and $r_h(s_{H+2},\ba)=1$. For the transition probability function $\PP_h$, $s_{H+1}$ and $s_{H+2}$ are absorbing states, which will always stay at the same state, and for other state $s_h(1\leq h\leq H)$, we have
\begin{align}
    &\PP_h(s_{h+1}|s_h,\ab)=1-\delta-\langle\bmu_h,\ab\rangle,\notag\\
    &\PP_h(s_{H+2}|s_h,\ab)=\delta+\langle\bmu_h,\ab\rangle\notag,
\end{align}
where each $\bmu_h\in \{-\Delta,\Delta\}^{d}$ and $\delta=1/H$. Furthermore, these hard-to-learn MDPs can be represented as linear mixture MDPs with the following feature mapping $\bphi: \cS \times \cS  \times \cA \rightarrow \RR^d$ and vector $\btheta_h$:
\begin{align}
    &\bphi(s_{h+1}|s_h,\ab) = \big(\alpha(1-\delta),-\beta\ab\big), h\in[H],\notag\\
    &\bphi(s_{H+2}|s_h,\ab) = \big(\alpha\delta,\beta\ab\big), h\in[H],\notag\\
    &\bphi(s_{h+1}|s_h,\ab) = \big(\alpha,\zero\big), h\in[H],\notag\\
    &\bphi(s_{h+1}|s_h,\ab) = (0,\zero), h\in[H],\notag\\
    &\btheta_h=(1/\alpha,\bmu_h/\beta),h\in[H]\notag,
\end{align}
where $\zero=\{0\}^{d-1}$ is a $(d-1)$-dimensional vector of all zeros, $\alpha=\sqrt{1/\big(1+(d-1)\Delta\big)}$ and $\beta=\sqrt{\Delta/\big(1+(d-1)\Delta\big)}$.
According to previous analysis on these hard-to-learn MDPs in \citet{zhou2020nearly}, we know that the regret of this MDP instance can be lower bounded by the regret of $H/2$ bandit instances. Thus, we will give a privatized version of Lemma C.7 in \citet{zhou2020nearly}.

Inspired by Lemma 3 in \citet{basu2019differential}, we first present the following locally differentially private KL-divergence decomposition lemma for $\varepsilon$-LDP contextual linear bandits.
% \qg{need to define $D(||)$.}
% \begin{lemma}\label{KLdecom}
%     (locally differentially private KL-divergence Decomposition) We denote the reward generated by user $t$ for action $\xb_t$ as $y_t = \xb_t^\top \btheta + \eta_t$, where $\eta_t$ is a zero-mean noise, and the observed history in the first $T$ steps as $ \cH_T = \cbr{\xb_i, y_t}_{i = 1}^T$. Then we have
%     \[\PP^T_{\pi\btheta} = \PP_{\pi\btheta} = \prod^T_{t=1}\pi (\xb_i\given \cH_{t-1})f_{\xb_i}(y_t)\,,\]
%     where $\pi$ is a bandit algorithm and $f_{\xb_i}$ is the distribution of reward with respect to action $\xb_t$, which is conditionally independent of the previous observed history $\cH_{t-1}$. If the reward generation process is $\varepsilon$-local differentially private for both the bandits with parameters $\btheta_1$ and $\btheta_2$, we have,
%     \[\text{KL}(\PP_{\pi\btheta_1}^T,\PP_{\pi\btheta_2}^T)\le 2\min\cbr{4, e^{2\varepsilon}}(e^{\varepsilon}-1)^2\sum_{t=1}^T \EE_{\pi\btheta_1}\sbr{\text{KL}(f_t^1(Z_t),f_t^2(Z_t))}\,,\]
%     where $Z_i$ is the privatized version of $y_t$.
% \end{lemma}

\begin{proof}[Proof of Lemma \ref{KLdecom}]
Instead of only protecting the output rewards in MAB algorithms with LDP guarantee, LDP contextual linear bandit algorithms requires the input information $\xb_t, y_t$ to the server to be protected by the privacy-preserving mechanism $\cM$. We denote by $f_{\cM}^{\btheta_1}, f_{\cM}^{\btheta_2}$ the privatized conditional distribution of reward given $\xb_t$, and $\tilde{\xb}_t, \tilde{y}_t$ are the privatized version of $\xb_t, y_t$, respectively. Combining the definition of ``the distribution of observed history''  $\PP_{\pi, \btheta}^T$ in \eqref{eq:definition of P} and  Lemma 3 in \citet{basu2019differential}, we have
\begin{align*}
    \text{KL}(\PP_{\pi, \btheta_1}^T,\PP_{\pi, \btheta_2}^T) &= \sum_{t=1}^T\EE_{\pi, \btheta_1}\sbr{\text{KL}(\cM_{\pi}(\tilde{\xb}_t\given \tilde{\cH}_{t-1})),\cM_{\pi}(\tilde{\xb}_t\given \tilde{\cH}_{t-1})))}\notag\\
    &\phantom{=\qquad} +\sum_{t=1}^T \EE_{\pi, \btheta_1}\sbr{\text{KL}(f^{\cM}_{\btheta_1}(\tilde{y}_t \given {\xb}_t), f^{\cM}_{\btheta_2}(\tilde{y}_t\given {\xb}_t))}\\
    &\le \sum_{t=1}^T \EE_{\pi, \btheta_1}\sbr{\text{KL}(f^{\cM}_{\btheta_1}(\tilde{y}_t \given {\xb}_t), f^{\cM}_{\btheta_2}(\tilde{y}_t\given {\xb}_t))}+\sum_{t=1}^T \EE_{\pi, \btheta_1}\sbr{\text{KL}(f^{\cM}_{\btheta_2}(\tilde{y}_t \given {\xb}_t), f^{\cM}_{\btheta_1}(\tilde{y}_t\given {\xb}_t))}\\
    &\le \min\cbr{4, e^{2\varepsilon}}(e^{\varepsilon}-1)^2\sum_{t=1}^T \EE_{\pi, \btheta_1}\nbr{f_{\btheta_1}(\tilde{y}_t \given {\xb}_t)-f_{\btheta_2}(\tilde{y}_t \given {\xb}_t)}_{\text{TV}}^2\\
    &\le 2\min\cbr{4, e^{2\varepsilon}}(e^{\varepsilon}-1)^2\sum_{t=1}^T \EE_{\pi, \btheta_1}\sbr{\text{KL}(f_{\btheta_1}(\tilde{y}_t \given {\xb}_t),f_{\btheta_2}(\tilde{y}_t \given {\xb}_t))}\,,
\end{align*}
%$f_{\cM}^{\btheta_2}, f_{\btheta_2}^{\cM}, f_{\btheta_2,\cM}$
where the first inequality is from the fact that KL divergence is non-negative. The second is obtained from Theorem 1 in \citet{duchi2018minimax} and the last inequality is due to Pinsker's inequality \citep{cover1999elements}. Therefore, we complete the proof of Lemma \ref{KLdecom} and obtain a result that is similar to Lemma 4 in \citet{basu2019differential}. 
\end{proof}
\subsection{Proof of Corollary \ref{banditLowerBound}}
Now, we begin the proof of Corollary \ref{banditLowerBound}, which give a lower bound for the regret of $\varepsilon$-LDP contextual linear bandits.
\begin{proof}[Proof of Corollary \ref{banditLowerBound}]
In this proof, we adapted the hypercube action set in \citet{lattimore2020bandit} (Theorem 24.1). Let the action set $\cA = \cbr{-1, 1}^d$ and $\bTheta = \cbr{-\Delta, \Delta}^d$. Let's define $\xb_t\in \cA$ to be the action chosen at step $t$. Given $\btheta \in \bTheta$, we have
\begin{align*}
    \text{Regret}(T) &= \Delta\EE_{\btheta}\sbr{\sum_{t=1}^T\sum_{i=1}^d (\sign(\theta_i) - \sign(x_{ti}))\sign(\theta_i)}\\
    &\ge \Delta\sum_{i=1}^d\EE_{\btheta} \sbr{\sum_{t=1}^T\mathds{1}\cbr{\sign(x_{ti})\ne \sign(\theta_{i})}}\\
    &\ge \Delta \frac{T}{2}\sum_{i=1}^d\PP_{\btheta} \rbr{\sum_{t=1}^T\mathds{1}\cbr{\sign(x_{ti})\ne \sign(\theta_{i})}\ge T/2}\,,
\end{align*}
where the first inequality holds due to the fact that $ (\sign(\theta_i) - \sign(x_{ti}))\sign(\theta_i)\ge \mathds{1}\cbr{\sign(x_{ti})\ne \sign(\theta_{i})} $ and the last inequality holds due to the Markov’s inequality.
For any vector $\btheta \in \bTheta$ and $i \in [d]$, we consider the vector $\btheta^{\prime} \in \bTheta$ such that $\theta^{\prime}_j = \theta_j$ for $j\ne i$ and $\theta^{\prime}_i \ne \theta_i$. We denote the event $\cE_{\btheta}$ as $\cE_{\btheta}=\cbr{\sum_{t=1}^T\mathds{1}\cbr{\sign(x_{ti})\ne \sign(\theta_{i})}\ge T/2}$. Then, according to the Bretagnolle-Huber inequality \citep{bretagnolle1979estimation} with the notation $p_{\theta_i} = \PP_{\btheta}(\cE_{\btheta})$ and $p_{\theta_i^{\prime}} = \PP_{\btheta^{\prime}}(\cE_{\btheta^{\prime}})$, we have
\begin{equation}
    p_{\theta_i}+p_{\theta_i^{\prime}}\ge \frac{1}{2}\exp(-\text{KL}(\PP_{\btheta},\PP_{\btheta^{\prime}}))\label{sump}\,.
\end{equation}
Suppose that $\delta\le 1/3, d\Delta\le \delta/2$, by Lemma \ref{KLdecom}, we further have
\begin{align}
    \text{KL}(\PP_{\btheta},\PP_{\btheta^{\prime}})&\le 2\min\cbr{4, e^{2\varepsilon}}(e^{\varepsilon}-1)^2\sum_{t=1}^T \EE_{\btheta}\sbr{\text{KL}(f_{\btheta}(\tilde{y}_t\given \xb_t),f_{\btheta^\prime}(\tilde{y}_t\given \xb_t))}\notag\\
    &= 2\min\cbr{4, e^{2\varepsilon}}(e^{\varepsilon}-1)^2\sum_{t=1}^T \EE_{\btheta}\sbr{\text{KL}\rbr{B(\left<\xb_t,\btheta\right>+\delta),B\rbr{\left<\xb_t,\btheta^{\prime}\right>+\delta}}}\notag\\
    &\le 2\min\cbr{4, e^{2\varepsilon}}(e^{\varepsilon}-1)^2 \sum_{t=1}^T \EE_{\btheta} \sbr{\frac{\left<\xb_t,\btheta-\btheta^{\prime}\right>^2}{\left<\xb_t,\btheta\right>+\delta}}\notag\\
    &\le 32\min\cbr{4, e^{2\varepsilon}}(e^{\varepsilon}-1)^2\frac{T\Delta^2}{\delta}\,,\label{KL}
\end{align}
where the first inequality holds due to Lemma \ref{KLdecom}, the second equality holds since for any two Bernoulli $B(a)$ and $B(b)$, we have $\text{KL}(B(a),B(b))\le 2(a-b)^2/a$ when $a\le 1/2, a+b\le 1$, and the last inequality holds due to the definition of vector $\btheta^{\prime}$ with the fact that $\left<\xb_t,\btheta\right>\ge -d\Delta\ge -\delta/2$.
Therefore, combining \eqref{sump} and \eqref{KL}, we further derive that
\[p_{\theta_i}+p_{\theta_i^{\prime}}\ge \frac{1}{2} \exp\rbr{-32\min\cbr{4, e^{2\varepsilon}}(e^{\varepsilon}-1)^2\frac{T\Delta^2}{\delta}}\,.\]
Taking average over all $\btheta \in \bTheta$, we get
\[\frac{1}{\abr{\bTheta}}\sum_{\btheta\in \bTheta}\sum_{i=1}^d p_{\theta_i}\ge \frac{d}{4}\exp\rbr{-32\min\cbr{4, e^{2\varepsilon}}(e^{\varepsilon}-1)^2\frac{T\Delta^2}{\delta}}\,.\]
Thus, there exists a $\btheta\in \bTheta$ such that $\sum_{i=1}^d p_{\theta_i}\ge d\exp\rbr{-32\min\cbr{4, e^{2\varepsilon}}(e^{\varepsilon}-1)^2T\Delta^2/\delta}/4$. By choosing 
$\Delta = \sqrt{\delta}/\rbr{\min\cbr{2, e^{\varepsilon}}(e^{\varepsilon}-1)\sqrt{T}}$, we obtain:
\[\text{Regret}(T)\ge \frac{\exp(-32)}{8\min\cbr{2, e^{\varepsilon}}(e^{\varepsilon}-1)}d\sqrt{T\delta}\,.\]
\end{proof}

\subsection{Proof of Theorem \ref{RLlower}}
\begin{proof}[Proof of Theorem \ref{RLlower}]
Our proof is based on the hard-to-learn instance $M(\cS, \cA, H, \cbr{r_h}, \cbr{\PP_h})$ constructed in \citet{zhou2020nearly}, where $\cA = \cbr{-1,1}^{d-1}, \delta = 1/H$ and $\bmu_h \in \cbr{-\Delta, \Delta}^{d-1}$. The only difference is that we set $\Delta = \sqrt{\delta}/\rbr{\min\cbr{2, e^{\varepsilon}}(e^{\varepsilon}-1)\sqrt{K}}$. According to Definition \ref{definMDP}, we choose $\bphi(s^\prime\given s, \ab), \btheta_h\in \RR^{d+1}$ as follows:
\begin{align*}
\bphi\rbr{s^{\prime} \given s, \mathbf{a}}=
\left\{\begin{array}{ll}
(\alpha(1-\delta),-\beta \mathbf{a}^{\top})^{\top}, & s=x_{h}, s^{\prime}=x_{h+1}, h \in[H] \\
(\alpha \delta, \beta \mathbf{a}^{\top})^{\top}, & s=x_{h}, s^{\prime}=x_{H+2}, h \in[H] ; \\
(\alpha, \mathbf{0}^{\top})^{\top}, & s \in\left\{x_{H+1}, x_{H+2}\right\}, s^{\prime}=s ; \\
\mathbf{0}, & \text {otherwise }\,.
\end{array}, \boldsymbol{\theta}_{h}=\left(1 / \alpha, \boldsymbol{\mu}_{h}^{\top} / \beta\right)^{\top}, h \in[H]\,.\right.
\end{align*}
% \begin{equation*}
%     \bphi(s,\ab)=
%     \begin{cases}
%     \rbr{\alpha, \beta\ab^{\top},0}^{\top}, &s= x_h;\\
%     \rbr{0, \mathbf{0}^{\top}, 1}, &s= x_{H+2}
%     \end{cases}, 
%     \bmu(s^{\prime}) = 
%     \begin{cases}
%     \rbr{(1-\delta)/\alpha, -\bmu_h^{\top}/\beta,0}^{\top}, &s= x_{h+1};\\
%     \rbr{\delta/\alpha, \bmu_h^{\top}/\beta, 1}, &s= x_{H+2};\\
%     \textbf{0}, &\text{otherwise}\,,
%     \end{cases}
% \end{equation*}
where $\alpha = \sqrt{1/(1+\Delta(d-1))}$ and $ \beta = \sqrt{\Delta/(1+\Delta(d-1))}$. Therefore, from Lemma C.7 in \citet{zhou2020nearly}, we directly obtain
\begin{align*}
    \sup_{\bmu}\EE_{\bmu} \text{Regret}\rbr{M_{\bmu}, K}&\ge\frac{H}{10}\sum_{h=1}^{H/2}\text{BanditRegret}(K)\\
    &\ge \frac{c }{\min\cbr{2, e^{\varepsilon}}(e^{\varepsilon}-1)}dH^2\sqrt{K\delta}\\
    &=\frac{c }{\min\cbr{2, e^{\varepsilon}}(e^{\varepsilon}-1)}dH\sqrt{T}\,,
\end{align*}
where $c$ is an absolute constant, the first inequality holds due to Lemma C.7 in \citet{zhou2020nearly} and  the last inequality follows by Corollary \ref{banditLowerBound}. Thus, we complete the proof of Theorem \ref{RLlower}.
\end{proof}
\section{Auxiliary Lemma}
% \subsection{Important inequlities for Wishart distribution}
% \begin{lemma}\label{eigenWishart}
%     (Lemma A.3, \citealt{sheffet2015private}). Fix $\alpha \in (0,1/e)$ and let $\Wb\sim \cW_d(\Vb, \nu)$ with $\sqrt{\nu}> \sqrt{d}+\sqrt{2\log(2/\alpha)}$, with probability at least $1-\alpha$ it holds that $\forall j=1,2,\dots, d$:
%     \[\sigma_j(\Wb)\in \rbr{\sqrt{\nu}\pm (\sqrt{d}+\sqrt{2 \log(2/\alpha)})}^2\sigma_j(\Vb)\,.\]
% \end{lemma}
% \begin{lemma}\label{wishartTail}
%     (Lemma A.1, \citealt{sheffet2015private}). Let $\Xb$ be a $(r\times d)$-matrix of $i.i.d$ normal Gaussians (i.e., $x_{ij} \sim \mathcal{N}(0,1) $). Fix $\beta\in(0,\frac{1}{e})$. Then, for any vector $\vb$ it holds that
%     \[\PP\sbr{\vb^\top\rbr{\frac{1}{r}\Xb\Xb^\top-\Ib}\vb\le\rbr{2\sqrt{2\frac{\log(2/\beta)}{r}}+2\frac{\log(2/\beta)}{r}}\nbr{\vb}^2}\ge 1-\beta\,.\]
    
%     Furthermore, if $r\ge d$ then denote $t=\sqrt{\frac{2\log(2/\beta)}{r-d+1}}$ and assume $t<1$. Then
%     \[\PP \sbr{\abr{\vb^{\top}\rbr{\Ib-\rbr{\frac{1}{r-d+1} \Xb^{\top}\Xb}^{-1}}\vb} \leq \frac{2t-t^2}{(1-t)^2} \nbr{\vb}^2 } \geq 1-\beta\,.\]
% \end{lemma}
%\subsection{Important inequality for summations}
\begin{lemma}\label{hoeffding}
    (Corollary 7, \citealt{jin2019short}).
    There exists an absolute constant c, such that if random vectors $\Xb_1, \dots, \Xb_n \in \RR^d$, and corresponding filtrations $\cF_i = \sigma(\Xb_1, \dots, \Xb_i)$ for $i\in [n]$ satisfy that $\Xb_i\given\cF_{i-1}$ is zero-mean $\sigma_i$-sub-Gaussian with fixed $\{\sigma_i\}$, then for any $\delta > 0$, with probability at least $1 - \delta$
    \[\nbr{\sum_{i=1}^n \Xb_i}_2 \le c\cdot\sqrt{\sum_{i=1}^n \sigma_i^2 \log \frac{2d}{\delta}}\,.\]
\end{lemma}
\begin{lemma}\label{boundPhi}
    (Lemma 11, \citealt{abbasi2011improved}). Let $\cbr{\bphi_t}_{t\ge 0}$ be a bounded sequence in $\RR_d$ satisfying $\sup_{t\ge 0}\nbr{\bphi_t}\le 1$. Let $\bLambda_0 \in \RR^{d\times d}$ be a positive definite matrix. For any $t \ge 0$, we define $\bLambda_t = \bLambda_0+\sum_{j=0}^t\bphi_j^\top\bphi_j$. Then, if the smallest eigenvalue of $\bLambda_0$ satisfies $\lambda_{\min}(\bLambda)\ge 1$, we have
    \[\log\sbr{\frac{\det (\bLambda_t)}{\det (\bLambda_0)}}\le \sum_{j=1}^t \bphi_j^\top \bLambda_{j-1}^{-1}\bphi_j\le 2 \log\sbr{\frac{\det(\bLambda_t)}{\det (\bLambda_0)}}\,.\]
\end{lemma}
% \subsection{Concentration inequalities for self-normalized processes}
% \begin{lemma}\label{covering}
%     (Lemma D.6, \citealt{jin2020provably}) 
%     Let $\mathcal{V}$ denote a class of functions mapping from $\cS$ to $\RR$ with following parametric form
%     \[V(\cdot)=\min \left\{\max_a r(\cdot, a) + \left<{\btheta}, \bphi(\cdot, a)\right> +\beta \nbr{\bSigma^{-1/2}\bphi (\cdot, a)}_2,H\right\}\,.\]

%     where the parameters $(\btheta, \beta, \bSigma)$ satisfy $\nbr{\btheta}_{2} \leq L, \beta \in [0,B]$ and the minimum eigenvalue satisfies $\rho_{\min }(\bSigma) \geq \lambda$. Assume $\nbr{\bphi(s,a)}_{2} \leq \sqrt{d}H$ for all $(s,a )$ pairs, and let $\mathcal{N}_{\varepsilon}$ be the $\varepsilon$-covering number of $\mathcal{V}$ with respect to the distance $\text{dist}(V,V^\prime)=\sup_x \left\vert V(s) - V^\prime (s) \right\vert $. Then
%     \[\log \mathcal{N}_\varepsilon \leq d \log (1+4L\sqrt{d}H/\varepsilon)+d^2 \log \sbr{1+8d^{5/2}H^2B^2/(\lambda \varepsilon^2)}\,.\]
% \end{lemma}
% % \begin{proof}
% % We adapt the proof in Lemma D.6, \citealt{jin2020provably}. For any two functions $V_1, V_2\in \cV$, we have
% % \begin{align*}
% %     \text{dist}(V_1, V_2) &\le \sup_{s, a} \nbr{\sbr{r(s, a) + \left<{\btheta}_1, \bphi_1(s, a)\right> +\beta \nbr{\bSigma_1^{-1/2}\bphi_1 (s, a)}_2} \\
% %     &\qquad- \sbr{r(s, a) + \left<{\btheta}, \bphi_2(s, a)\right> +\beta \nbr{\bSigma_2^{-1/2}\bphi (\cdot, a)}_2}}\\
% %     &
% % \end{align*}
% % \end{proof}

%\subsection{Proof of Lemma \ref{LEMMA:TRANSITION1}}

\bibliographystyle{ims}
\bibliography{reference}

\end{document}